\documentclass[english]{article}
\usepackage[T1]{fontenc}
\usepackage[latin9]{inputenc}
\usepackage{amsmath}
\usepackage{graphicx}
\usepackage{esint}
\usepackage{babel}

\begin{document}

\title{Self-Organising Stochastic Encoders%
\thanks{This talk was given at the Workshop on Self-Organising Systems - Future
Prospects for Computing, 28-29 October 1999, Manchester, UK.%
}}

\author{S P Luttrell}
\maketitle
\begin{abstract}
The processing of mega-dimensional data, such as images, scales linearly
with image size only if fixed size processing windows are used. It
would be very useful to be able to automate the process of sizing
and interconnecting the processing windows. A stochastic encoder that
is an extension of the standard Linde-Buzo-Gray vector quantiser,
called a stochastic vector quantiser (SVQ), includes this required
behaviour amongst its emergent properties, because it automatically
splits the input space into statistically independent subspaces, which
it then separately encodes.

Various optimal SVQs have been obtained, both analytically and numerically.
Analytic solutions which demonstrate how the input space is split
into independent subspaces may be obtained when an SVQ is used to
encode data that lives on a 2-torus (e.g. the superposition of a pair
of uncorrelated sinusoids). Many numerical solutions have also been
obtained, using both SVQs and chains of linked SVQs: (1) images of
multiple independent targets (encoders for single targets emerge),
(2) images of multiple correlated targets (various types of encoder
for single and multiple targets emerge), (3) superpositions of various
waveforms (encoders for the separate waveforms emerge - this is a
type of independent component analysis (ICA)), (4) maternal and foetal
ECGs (another example of ICA), (5) images of textures (orientation
maps and dominance stripes emerge).

Overall, SVQs exhibit a rich variety of self-organising behaviour,
which effectively discovers the internal structure of the training
data. This should have an immediate impact on {}``intelligent''
computation, because it reduces the need for expert human intervention
in the design of data processing algorithms.
\end{abstract}

\section{Stochastic Vector Quantiser}

\subsection{Reference}

\noindent Luttrell S P, 1997, \emph{Mathematics of Neural Networks:
Models, Algorithms and Applications}, Kluwer, Ellacott S W, Mason
J C and Anderson I J (eds.), A theory of self-organising neural networks,
240-244.

\subsection{Objective Function}

\subsubsection{Mean Euclidean Distortion}

\begin{equation}
D=\int d\boldsymbol{x}\,\Pr\left(\boldsymbol{x}\right)\sum_{\boldsymbol{y}}\Pr\left(\boldsymbol{y}\left|\boldsymbol{x}\right.\right)\int d\boldsymbol{x}^{\prime}\Pr\left(\boldsymbol{x}^{\prime}\left|\boldsymbol{y}\right.\right)\left\Vert \boldsymbol{x}-\boldsymbol{x}^{\prime}\right\Vert ^{2}\end{equation}

\begin{itemize}
\item Encode then decode: $\boldsymbol{x}\longrightarrow\boldsymbol{y}\longrightarrow\boldsymbol{x}^{\prime}$.
\item $\boldsymbol{x}$ = input vector; $\boldsymbol{y}$ = code; $\boldsymbol{x}^{\prime}$
= reconstructed vector.
\item Code vector $\boldsymbol{y}=\left(y_{1},y_{2},\,\cdots\,,y_{n}\right)$
for $1\le y_{i}\le M$.
\item $\Pr\left(\boldsymbol{x}\right)$ = input PDF; $\Pr\left(\boldsymbol{y}\left|\boldsymbol{x}\right.\right)$
= stochastic encoder; $\Pr\left(\boldsymbol{x}^{\prime}\left|\boldsymbol{y}\right.\right)$
= stochastic decoder.
\item $\left\Vert \boldsymbol{x}-\boldsymbol{x}^{\prime}\right\Vert ^{2}$
= Euclidean reconstruction error.
\end{itemize}

\subsubsection{Simplify}

\begin{eqnarray}
D & = & 2\int d\boldsymbol{x}\,\Pr\left(\boldsymbol{x}\right)\sum_{\boldsymbol{y}}\Pr\left(\boldsymbol{y}\left|\boldsymbol{x}\right.\right)\left\Vert \boldsymbol{x}-\boldsymbol{x}^{\prime}\left(\boldsymbol{y}\right)\right\Vert ^{2}\nonumber \\
\boldsymbol{x}^{\prime}\left(\boldsymbol{y}\right) & \equiv & \int d\boldsymbol{x}\,\Pr\left(\boldsymbol{x}\left|\boldsymbol{y}\right.\right)\boldsymbol{x}=\frac{\int d\boldsymbol{x}\,\Pr\left(\boldsymbol{x}\right)\Pr\left(\boldsymbol{y}\left|\boldsymbol{x}\right.\right)\boldsymbol{x}}{\int d\boldsymbol{x}\,\Pr\left(\boldsymbol{x}\right)\Pr\left(\boldsymbol{y}\left|\boldsymbol{x}\right.\right)}\end{eqnarray}

\begin{itemize}
\item Do the $\int d\boldsymbol{x}\Pr\left(\boldsymbol{x}\left|\boldsymbol{y}\right.\right)\left(\cdots\right)$
integration.
\item $\boldsymbol{x}^{\prime}\left(\boldsymbol{y}\right)$ = reconstruction
vector.
\item $\boldsymbol{x}^{\prime}\left(\boldsymbol{y}\right)$ is the solution
of $\frac{\partial D}{\partial\boldsymbol{x}^{\prime}\left(\boldsymbol{y}\right)}=0$,
so it can be deduced by optimisation.
\end{itemize}

\subsubsection{Constrain}

\begin{eqnarray}
\Pr\left(\boldsymbol{y}\left|\boldsymbol{x}\right.\right) & = & \Pr\left(y_{1}\left|\boldsymbol{x}\right.\right)\Pr\left(y_{2}\left|\boldsymbol{x}\right.\right)\,\cdots\,\Pr\left(y_{n}\left|\boldsymbol{x}\right.\right)\nonumber \\
\boldsymbol{x}^{\prime}\left(\boldsymbol{y}\right) & = & \frac{1}{n}\sum_{i=1}^{n}\boldsymbol{x}^{\prime}\left(y_{i}\right)\end{eqnarray}

\begin{itemize}
\item $\Pr\left(\boldsymbol{y}\left|\boldsymbol{x}\right.\right)$ implies
the components $\left(y_{1},y_{2},\,\cdots\,,y_{n}\right)$ of $\boldsymbol{y}$
are conditionally independent given $\boldsymbol{x}$.
\item $\boldsymbol{x}^{\prime}\left(\boldsymbol{y}\right)$ implies the
reconstruction is a superposition of contributions $\boldsymbol{x}^{\prime}\left(y_{i}\right)$
for $i=1,2,\,\cdots\,,n$. 
\item The stochastic encoder samples $n$ times from the same $\Pr\left(y\left|\boldsymbol{x}\right.\right)$.
\end{itemize}

\subsubsection{Upper Bound}

\begin{eqnarray}
D & \le & D_{1}+D_{2}\nonumber \\
D_{1} & \equiv & \frac{2}{n}\int d\boldsymbol{x}\,\Pr\left(\boldsymbol{x}\right)\sum_{y=1}^{M}\Pr\left(y\left|\boldsymbol{x}\right.\right)\left\Vert \boldsymbol{x}-\boldsymbol{x}^{\prime}\left(y\right)\right\Vert ^{2}\nonumber \\
D_{2} & \equiv & \frac{2\left(n-1\right)}{n}\int d\boldsymbol{x}\,\Pr\left(\boldsymbol{x}\right)\left\Vert \boldsymbol{x}-\sum_{y=1}^{M}\Pr\left(y\left|\boldsymbol{x}\right.\right)\boldsymbol{x}^{\prime}\left(y\right)\right\Vert ^{2}\end{eqnarray}

\begin{itemize}
\item $D_{1}$ is a stochastic vector quantiser with the vector code $\boldsymbol{y}$
replaced by a scalar code $y$.
\item $D_{2}$ is a non-linear (note $\Pr\left(y\left|\boldsymbol{x}\right.\right)$)
encoder with a superposition term $\sum_{y=1}^{M}\Pr\left(y\left|\boldsymbol{x}\right.\right)\boldsymbol{x}^{\prime}\left(y\right)$.
\item $n\longrightarrow\infty$: the stochastic encoder measures $\Pr\left(y\left|\boldsymbol{x}\right.\right)$
accurately and $D_{2}$ dominates.
\item $n\longrightarrow1$: the stochastic encoder samples $\Pr\left(y\left|\boldsymbol{x}\right.\right)$
poorly and $D_{1}$ dominates. 
\end{itemize}

\section{Analytic Optimisation}

\subsection{References}

\noindent Luttrell S P, 1999, \emph{Combining Artificial Neural Nets:
Ensemble and Modular Multi-Net Systems}, 235-263, Springer-Verlag,
Sharkey A J C (ed.), Self-organised modular neural networks for encoding
data. \\

\noindent Luttrell S P, 1999, An Adaptive Network For Encoding Data
Using Piecewise Linear Functions, \emph{Proceedings of the 9th International
Conference on Artificial Neural Networks (ICANN99)}, Edinburgh, 7-10
September 1999, 198-203.

\subsection{Stationarity Conditions}

\subsubsection{Stationarity w.r.t. $\boldsymbol{x}^{\prime}\left(y\right)$}

\begin{equation}
n\int d\boldsymbol{x}\Pr\left(\boldsymbol{x}\left|y\right.\right)\boldsymbol{x}=\boldsymbol{x}^{\prime}\left(y\right)+\left(n-1\right)\int d\boldsymbol{x}\Pr\left(\boldsymbol{x}\left|y\right.\right)\sum_{y^{\prime}=1}^{M}\Pr\left(y^{\prime}\left|\boldsymbol{x}\right.\right)\boldsymbol{x}^{\prime}\left(y^{\prime}\right)\end{equation}

\begin{itemize}
\item Stationarity condition is $\frac{\partial\left(D_{1}+D_{2}\right)}{\partial\boldsymbol{x}^{\prime}\left(y\right)}=0$.
\end{itemize}

\subsubsection{Stationarity w.r.t. $\Pr\left(y^{\prime}\left|\boldsymbol{x}\right.\right)$}

\begin{equation}
\Pr\left(\boldsymbol{x}\right)\Pr\left(y\left|\boldsymbol{x}\right.\right)\sum_{y^{\prime}=1}^{M}\left(\Pr\left(y^{\prime}\left|\boldsymbol{x}\right.\right)-\delta_{y,y^{\prime}}\right)\boldsymbol{x}^{\prime}\left(y^{\prime}\right).\left(\frac{1}{2}\,\boldsymbol{x}^{\prime}\left(y^{\prime}\right)-n\,\boldsymbol{x}+\left(n-1\right)\sum_{y^{\prime\prime}=1}^{M}\Pr\left(y^{\prime\prime}\left|\boldsymbol{x}\right.\right)\boldsymbol{x}^{\prime}\left(y^{\prime\prime}\right)\right)=0\end{equation}

\begin{itemize}
\item Stationarity condition is $\frac{\partial\left(D_{1}+D_{2}\right)}{\partial\log\Pr\left(y\left|\boldsymbol{x}\right.\right)}=0$,
subject to $\sum_{y=1}^{M}\Pr\left(y\left|\boldsymbol{x}\right.\right)=1$.
\item 3 types of solution: $\Pr\left(\boldsymbol{x}\right)=0$ (trivial),
$\Pr\left(y\left|\boldsymbol{x}\right.\right)=0$ (ensures $\Pr\left(y\left|\boldsymbol{x}\right.\right)\ge0$),
and $\sum_{y^{\prime}=1}^{M}\left(\Pr\left(y^{\prime}\left|\boldsymbol{x}\right.\right)-\delta_{y,y^{\prime}}\right)\left(\cdots\right)=0$.
\end{itemize}

\subsection{Circle}

\subsubsection{Input vector uniformly distributed on a circle}

\begin{eqnarray}
\boldsymbol{x} & = & \left(\cos\theta,\sin\theta\right)\nonumber \\
\int d\boldsymbol{x}\,\Pr\left(\boldsymbol{x}\right)\left(\cdots\right) & = & \frac{1}{2\pi}\int_{0}^{2\pi}d\theta\left(\cdots\right)\end{eqnarray}

\subsubsection{Stochastic encoder PDFs symmetrically arranged around the circle}

\begin{equation}
\Pr\left(y\left|\theta\right.\right)=p\left(\theta-\frac{2\pi y}{M}\right)\end{equation}

\subsubsection{Reconstruction vectors symmetrically arraged around the circle}

\begin{equation}
\boldsymbol{x}^{\prime}\left(y\right)=r\,\left(\cos\left(\frac{2\pi y}{M}\right),\sin\left(\frac{2\pi y}{M}\right)\right)\end{equation}

\subsubsection{Stochastic encoder PDFs overlap no more than 2 at a time}

\begin{equation}
p\left(\theta\right)=\begin{cases}
1 & 0\le\left|\theta\right|\le\frac{\pi}{M}-s\\
f\left(\theta\right) & \frac{\pi}{M}-s\le\left|\theta\right|\le\frac{\pi}{M}+s\\
0 & \left|\theta\right|\ge\frac{\pi}{M}+s\end{cases}\end{equation}

\begin{center}
\includegraphics[width=8cm]{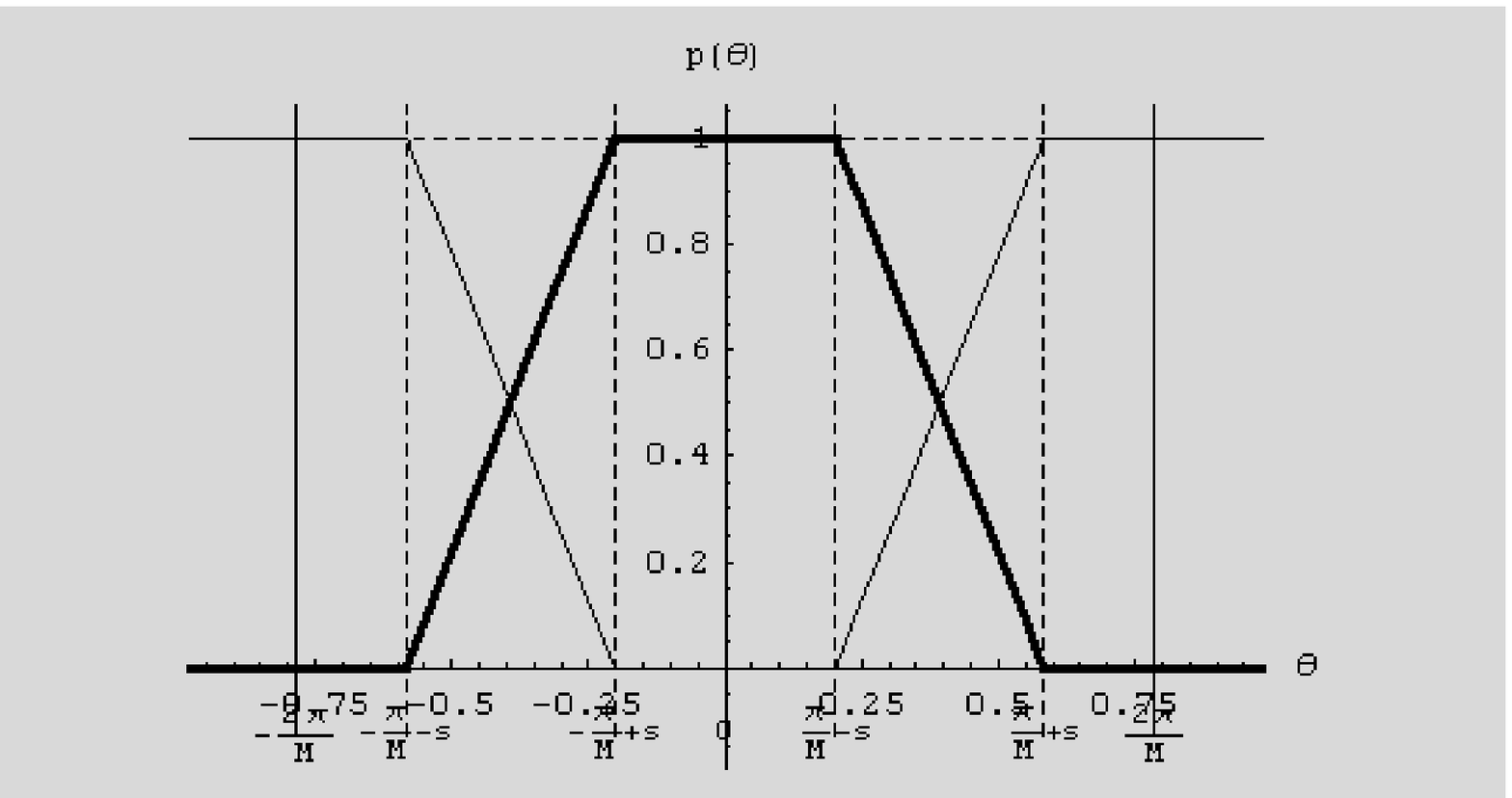}
\par\end{center}

\subsubsection{Stochastic encoder PDFs overlap no more than 3 at a time}

\begin{equation}
p\left(\theta\right)=\begin{cases}
f_{1}\left(\theta\right) & 0\le\left|\theta\right|\le-\frac{\pi}{M}+s\\
f_{2}\left(\theta\right) & -\frac{\pi}{M}+s\le\left|\theta\right|\le\frac{3\pi}{M}-s\\
f_{3}\left(\theta\right) & \frac{3\pi}{M}-s\le\left|\theta\right|\le\frac{\pi}{M}+s\\
0 & \left|\theta\right|\ge\frac{\pi}{M}+s\end{cases}\end{equation}

\begin{center}
\includegraphics[width=8cm]{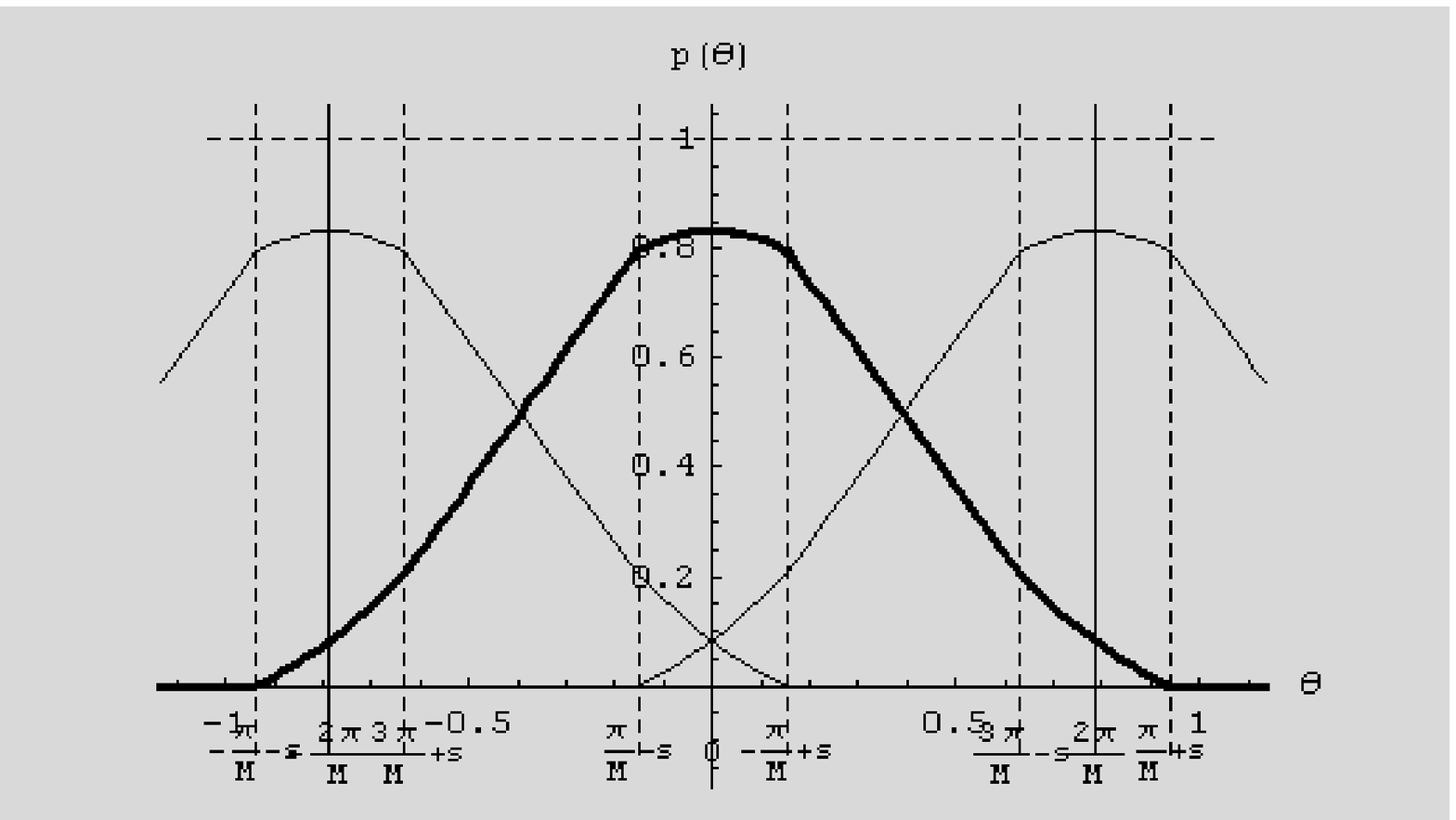}
\par\end{center}

\subsection{2-Torus}

\subsubsection{Input vector uniformly distributed on a 2-torus}

\begin{eqnarray}
\boldsymbol{x} & = & \left(\boldsymbol{x}_{1},\boldsymbol{x}_{2}\right)\nonumber \\
\boldsymbol{x}_{1} & = & \left(\cos\theta_{1},\sin\theta_{1}\right)\nonumber \\
\boldsymbol{x}_{2} & = & \left(\cos\theta_{2},\sin\theta_{2}\right)\nonumber \\
\int d\boldsymbol{x}\,\Pr\left(\boldsymbol{x}\right)\left(\cdots\right) & = & \frac{1}{4\pi^{2}}\int_{0}^{2\pi}d\theta_{1}\int_{0}^{2\pi}d\theta_{2}\left(\cdots\right)\end{eqnarray}

\subsubsection{Joint encoding}

\begin{equation}
\Pr\left(y\left|\boldsymbol{x}\right.\right)=\Pr\left(y\left|\boldsymbol{x}_{1},\boldsymbol{x}_{2}\right.\right)\end{equation}

\begin{itemize}
\item $\Pr\left(y\left|\boldsymbol{x}\right.\right)$ depends jointly on
$\boldsymbol{x}_{1}$ and $\boldsymbol{x}_{2}$.
\item Requires $n=1$ to encode $\boldsymbol{x}$.
\item For a given resolution the size of the codebook increases exponentially
with input dimension.
\end{itemize}
\begin{center}
\includegraphics[width=8cm]{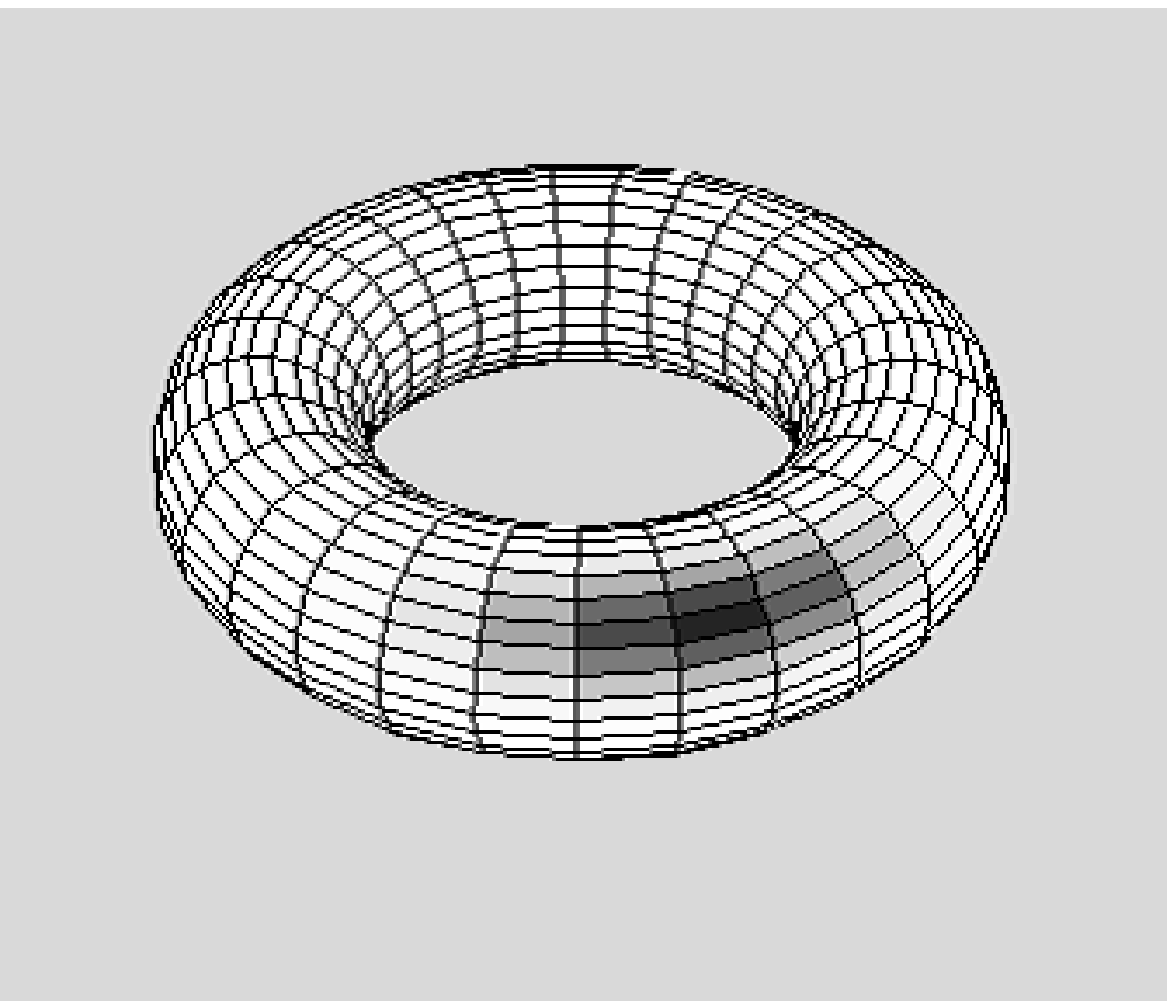}
\par\end{center}

\subsubsection{Factorial encoding}

\begin{equation}
\Pr\left(y\left|\boldsymbol{x}\right.\right)=\begin{cases}
\Pr\left(y\left|\boldsymbol{x}_{1}\right.\right) & y\in Y_{1}\\
\Pr\left(y\left|\boldsymbol{x}_{2}\right.\right) & y\in Y_{2}\end{cases}\end{equation}

\begin{itemize}
\item $Y_{1}$ and $Y_{2}$ are non-intersecting subsets of the allowed
values of $y$.
\item $\Pr\left(y\left|\boldsymbol{x}\right.\right)$ depends either on
$\boldsymbol{x}_{1}$ or on $\boldsymbol{x}_{2}$, but not on both
at the same time.
\item Requires $n\gg1$ to encode $\boldsymbol{x}$.
\item For a given resolution the size of the codebook increases linearly
with input dimension.
\end{itemize}
\begin{center}
\includegraphics[width=8cm]{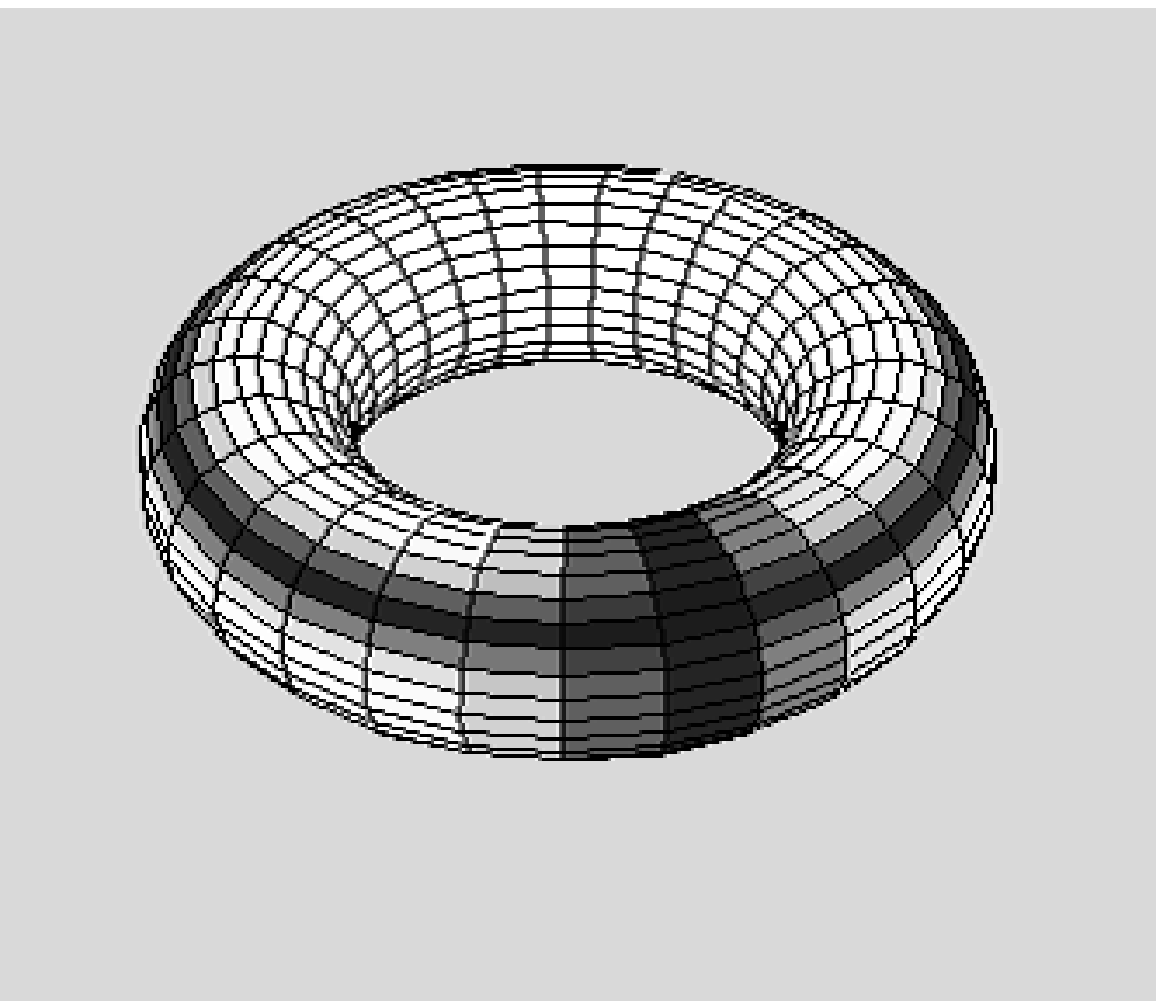}
\par\end{center}

\subsubsection{Stability diagram}
\begin{itemize}
\item Fixed $n$, increasing $M$: joint encoding is eventually favoured
because the size of the codebook is eventually large enough.
\item Fixed $M$, increase $n$: factorial encoding is eventually favoured
because the number of samples is eventually large enough.
\item Factorial encoding is encouraged by using a small codebook and sampling
a large number of times.
\end{itemize}
\begin{center}
\includegraphics[width=8cm]{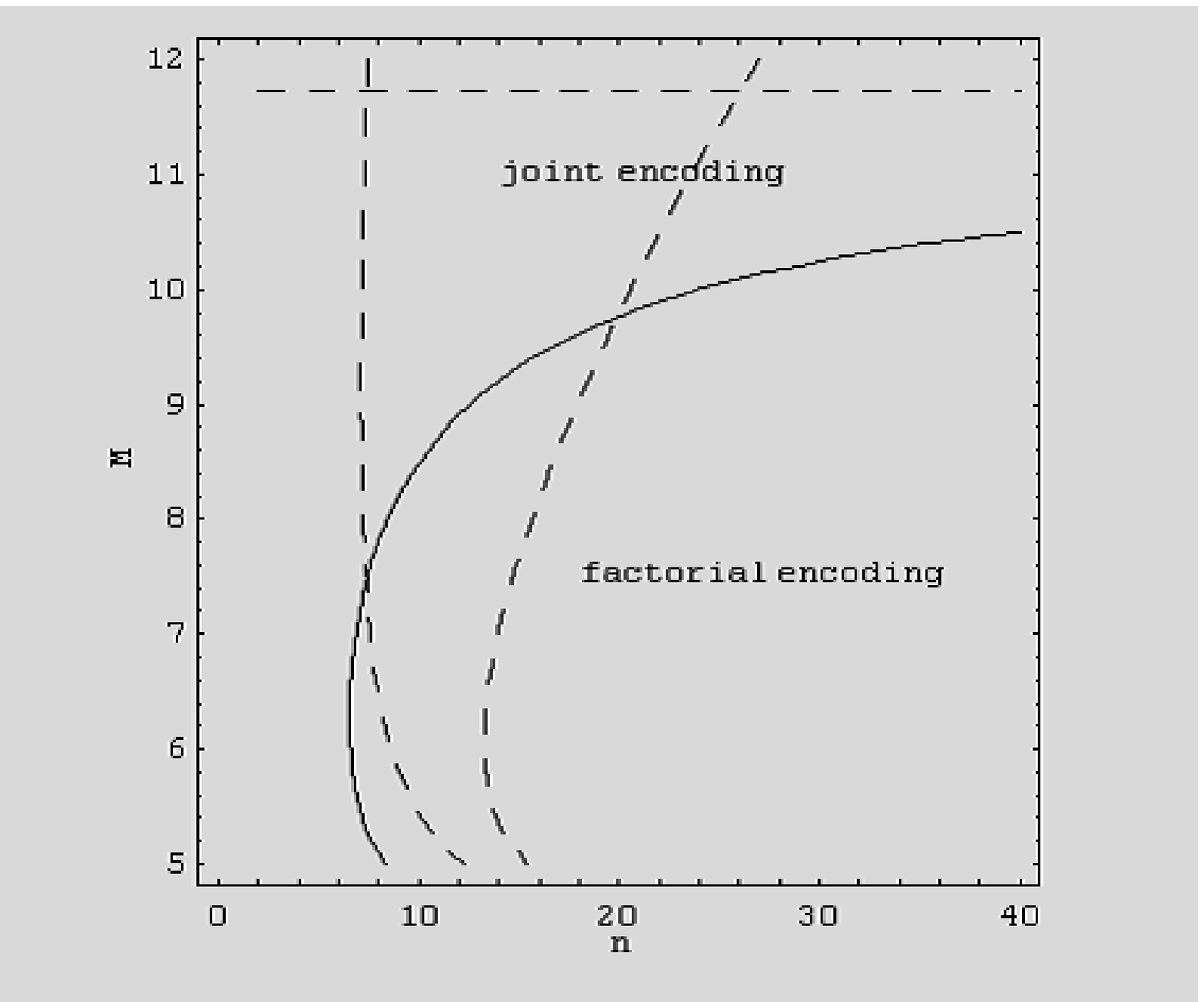}
\par\end{center}

\section{Numerical Optimisation}

\subsection{References}

\noindent Luttrell S P, 1997, to appear in \emph{Proceedings of the
Conference on Information Theory and the Brain}, Newquay, 20-21 September
1996, The emergence of dominance stripes and orientation maps in a
network of firing neurons. \\

\noindent Luttrell S P, 1997, \emph{Mathematics of Neural Networks:
Models, Algorithms and Applications}, Kluwer, Ellacott S W, Mason
J C and Anderson I J (eds.), A theory of self-organising neural networks,
240-244.\\

\noindent Luttrell S P, 1999, submitted to a special issue of \emph{IEEE
Trans. Information Theory on Information-Theoretic Imaging}, Stochastic
vector quantisers.

\subsection{Gradient Descent}

\subsubsection{Posterior probability with infinite range neighbourhood}

\[
\Pr\left(y\left|\boldsymbol{x}\right.\right)=\frac{Q\left(\boldsymbol{x}\left|y\right.\right)}{\sum_{y^{\prime}=1}^{M}Q\left(\boldsymbol{x}\left|y^{\prime}\right.\right)}\]

\begin{itemize}
\item $Q\left(\boldsymbol{x}\left|y\right.\right)\ge0$ is needed to ensure
a valid $\Pr\left(y\left|\boldsymbol{x}\right.\right)$.
\item This does not restrict $\Pr\left(y\left|\boldsymbol{x}\right.\right)$
in any way.
\end{itemize}

\subsubsection{Posterior probability with finite range neighbourhood}

\begin{equation}
\Pr\left(y\left|\boldsymbol{x};y^{\prime}\right.\right)\equiv\frac{Q\left(\boldsymbol{x}\left|y\right.\right)\delta_{y\in\mathcal{N}\left(y^{\prime}\right)}}{\sum_{y^{\prime\prime}=\mathcal{N}\left(y^{\prime}\right)}Q\left(\boldsymbol{x}\left|y^{\prime\prime}\right.\right)}\end{equation}

\begin{equation}
\Pr\left(y\left|\boldsymbol{x}\right.\right)=\frac{1}{M}\sum_{y^{\prime}=\mathcal{N}^{-1}\left(y\right)}\Pr\left(y\left|\boldsymbol{x};y^{\prime}\right.\right)=\frac{1}{M}\, Q\left(\boldsymbol{x}\left|y\right.\right)\sum_{y^{\prime}=\mathcal{N}^{-1}\left(y\right)}\frac{1}{\sum_{y^{\prime\prime}=\mathcal{N}\left(y^{\prime}\right)}Q\left(\boldsymbol{x}\left|y^{\prime\prime}\right.\right)}\end{equation}

\begin{itemize}
\item $\mathcal{N}\left(y^{\prime}\right)$ is the set of neurons that lie
in a predefined {}``neighbourhood'' of $y^{\prime}$.
\item $\mathcal{N}^{-1}\left(y\right)$ is the {}``inverse neighbourhood''
of $y$ defined as $\mathcal{N}^{-1}\left(y\right)\equiv\left\{ y^{\prime}:y\in\mathcal{N}\left(y^{\prime}\right)\right\} $.
\item Neighbourhood is used to introduce {}``lateral inhibition'' between
the firing neurons.
\item This restricts $\Pr\left(y\left|\boldsymbol{x}\right.\right)$, but
allows limited range lateral interactions to be used.
\end{itemize}

\subsubsection{Probability leakage}

\begin{equation}
\Pr\left(y\left|\boldsymbol{x}\right.\right)\longrightarrow\sum_{y^{\prime}\in\mathcal{L}^{-1}\left(y\right)}\Pr\left(y\left|y^{\prime}\right.\right)\Pr\left(y^{\prime}\left|\boldsymbol{x}\right.\right)\end{equation}

\begin{itemize}
\item $\Pr\left(y\left|y^{\prime}\right.\right)$ is the amount of probability
that leaks from location $y^{\prime}$ to location $y$.
\item $\mathcal{L}\left(y^{\prime}\right)$ is the {}``leakage neighbourhood''
of $y^{\prime}$.
\item $\mathcal{L}^{-1}\left(y\right)$ is the {}``inverse leakage neighbourhood''
of $y$ defined as $\mathcal{L}^{-1}\left(y\right)\equiv\left\{ y^{\prime}:y\in\mathcal{L}\left(y^{\prime}\right)\right\} $.
\item Leakage is to allow the network output to be {}``damaged'' in a
controlled way.
\item When the network is optimised it automatically becomes robust with
respect to such damage.
\item Leakage leads to topographic ordering according to the defined neighbourhood.
\item This restricts $\Pr\left(y\left|\boldsymbol{x}\right.\right)$, but
allows topographic ordering to be obtained, and is faster to train.
\end{itemize}

\subsubsection{Shorthand notation}

\begin{equation}
\begin{array}{ll}
L_{y,y^{\prime}}\equiv\Pr\left(y\left|y^{\prime}\right.\right) & P_{y,y^{\prime}}\equiv\Pr\left(y\left|\boldsymbol{x};y^{\prime}\right.\right)\\
p_{y}\equiv\sum_{y^{\prime}\in\mathcal{N}^{-1}\left(y\right)}P_{y^{\prime},y} & \left(L^{T}p\right)_{y}\equiv\sum_{y^{\prime}\in\mathcal{L}^{-1}\left(y\right)}L_{y^{\prime},y}\, p_{y^{\prime}}\\
\boldsymbol{d}_{y}\equiv\boldsymbol{x}-\boldsymbol{x}^{\prime}\left(y\right) & \left(L\,\boldsymbol{d}\right)_{y}\equiv\sum_{y^{\prime}\in\mathcal{L}\left(y\right)}L_{y^{\prime},y}\,\boldsymbol{d}_{y^{\prime}}\\
\left(P\, L\,\boldsymbol{d}\right)_{y}\equiv\sum_{y^{\prime}\in\mathcal{N}\left(y\right)}P_{y,y^{\prime}}\left(L\,\boldsymbol{d}\right)_{y^{\prime}} & \left(P^{T}P\, L\,\boldsymbol{d}\right)_{y}\equiv\sum_{y^{\prime}\in\mathcal{N}^{-1}\left(y\right)}P_{y^{\prime},y}\left(P\, L\,\boldsymbol{d}\right)_{y^{\prime}}\\
e_{y}\equiv\left\Vert \boldsymbol{x}-\boldsymbol{x}^{\prime}\left(y\right)\right\Vert ^{2} & \left(L\, e\right)_{y}\equiv\sum_{y^{\prime}\in\mathcal{L}\left(y\right)}L_{y,y^{\prime}}\, e_{y^{\prime}}\\
\left(P\, L\, e\right)_{y}\equiv\sum_{y^{\prime}\in\mathcal{N}\left(y\right)}P_{y,y^{\prime}}\left(L\, e\right)_{y^{\prime}} & \left(P^{T}P\, L\, e\right)_{y}\equiv\sum_{y^{\prime}\in N^{-1}\left(y\right)}P_{y^{\prime},y}\left(P\, L\, e\right)_{y^{\prime}}\\
\bar{\boldsymbol{d}}\equiv\sum_{y=1}^{M}\left(L^{T}p\right)_{y}\boldsymbol{d}_{y} & \mathrm{or\,}\bar{\boldsymbol{d}}\equiv\sum_{y=1}^{M}\left(P\, L\,\boldsymbol{d}\right)_{y}\end{array}\end{equation}

\begin{itemize}
\item This shorthand notation simplifies the appearance of the gradients
of $D_{1}$ and $D_{2}$.
\item For instance, $\Pr\left(y\left|\boldsymbol{x}\right.\right)=\frac{1}{M}\left(L^{T}p\right)_{y}$.
\end{itemize}

\subsubsection{Derivatives w.r.t. $\boldsymbol{x}^{\prime}\left(y\right)$}

\begin{eqnarray}
\frac{\partial D_{1}}{\boldsymbol{x}^{\prime}\left(y\right)} & = & -\frac{4}{n\, M}\int d\boldsymbol{x}\Pr\left(\boldsymbol{x}\right)\left(L^{T}p\right)_{y}\boldsymbol{d}_{y}\nonumber \\
\frac{\partial D_{2}}{\boldsymbol{x}^{\prime}\left(y\right)} & = & -\frac{4\left(n-1\right)}{n\, M^{2}}\int d\boldsymbol{x}\Pr\left(\boldsymbol{x}\right)\left(L^{T}p\right)_{y}\bar{\boldsymbol{d}}\end{eqnarray}

\begin{itemize}
\item The extra factor $\frac{1}{M}$ in $\frac{\partial D_{2}}{\boldsymbol{x}^{\prime}\left(y\right)}$
arises because there is a $\sum_{y=1}^{M}\left(\cdots\right)$ hidden
inside the $\bar{\boldsymbol{d}}$.
\end{itemize}

\subsubsection{Functional derivatives w.r.t. $\log Q\left(\boldsymbol{x}\left|y\right.\right)$}

\begin{eqnarray}
\delta D_{1} & = & \frac{2}{n\, M}\int d\boldsymbol{x}\Pr\left(\boldsymbol{x}\right)\sum_{y=1}^{M}\delta\log Q\left(\boldsymbol{x}\left|y\right.\right)\left(p_{y}\left(L\, e\right)_{y}-\left(P^{T}P\, L\, e\right)_{y}\right)\\
\delta D_{2} & = & \frac{4\left(n-1\right)}{n\, M^{2}}\int d\boldsymbol{x}\Pr\left(\boldsymbol{x}\right)\sum_{y=1}^{M}\delta\log Q\left(\boldsymbol{x}\left|y\right.\right)\left(p_{y}\left(L\,\boldsymbol{d}\right)_{y}-\left(P^{T}P\, L\,\boldsymbol{d}\right)_{y}\right).\bar{\boldsymbol{d}}\nonumber \end{eqnarray}

\begin{itemize}
\item Differentiate w.r.t. $\log Q\left(\boldsymbol{x}\left|y\right.\right)$
because $Q\left(\boldsymbol{x}\left|y\right.\right)\ge0$.
\end{itemize}

\subsubsection{Neural response model}

\begin{equation}
Q\left(\boldsymbol{x}\left|y\right.\right)=\frac{1}{1+\exp\left(-\boldsymbol{w}\left(y\right).\boldsymbol{x}-b\left(y\right)\right)}\end{equation}

\begin{itemize}
\item This is a standard {}``sigmoid'' function.
\item This restricts $\Pr\left(y\left|\boldsymbol{x}\right.\right)$, but
it is easy to implement, and leads to results similar to the ideal
analytic results.
\end{itemize}

\subsubsection{Derivatives w.r.t. $\boldsymbol{w}\left(y\right)$ and $b\left(y\right)$}

\begin{eqnarray}
\frac{\partial D_{1}}{\partial\left(\begin{array}{c}
b\left(y\right)\\
\boldsymbol{w}\left(y\right)\end{array}\right)} & = & \frac{2}{n\, M}\int d\boldsymbol{x}\Pr\left(\boldsymbol{x}\right)\left(p_{y}\left(L\, e\right)_{y}-\left(P^{T}P\, L\, e\right)_{y}\right)\left(1-Q\left(\boldsymbol{x}\left|y\right.\right)\right)\left(\begin{array}{c}
1\\
\boldsymbol{x}\end{array}\right)\\
\frac{\partial D_{2}}{\partial\left(\begin{array}{c}
b\left(y\right)\\
\boldsymbol{w}\left(y\right)\end{array}\right)} & = & \frac{4\left(n-1\right)}{n\, M^{2}}\int d\boldsymbol{x}\Pr\left(\boldsymbol{x}\right)\left(p_{y}\left(L\,\boldsymbol{d}\right)_{y}-\left(P^{T}P\, L\,\boldsymbol{d}\right)_{y}\right).\bar{\boldsymbol{d}}\left(1-Q\left(\boldsymbol{x}\left|y\right.\right)\right)\left(\begin{array}{c}
1\\
\boldsymbol{x}\end{array}\right)\nonumber \end{eqnarray}

\subsection{Circle}

\subsubsection{Training history}

\begin{center}
\includegraphics[width=8cm]{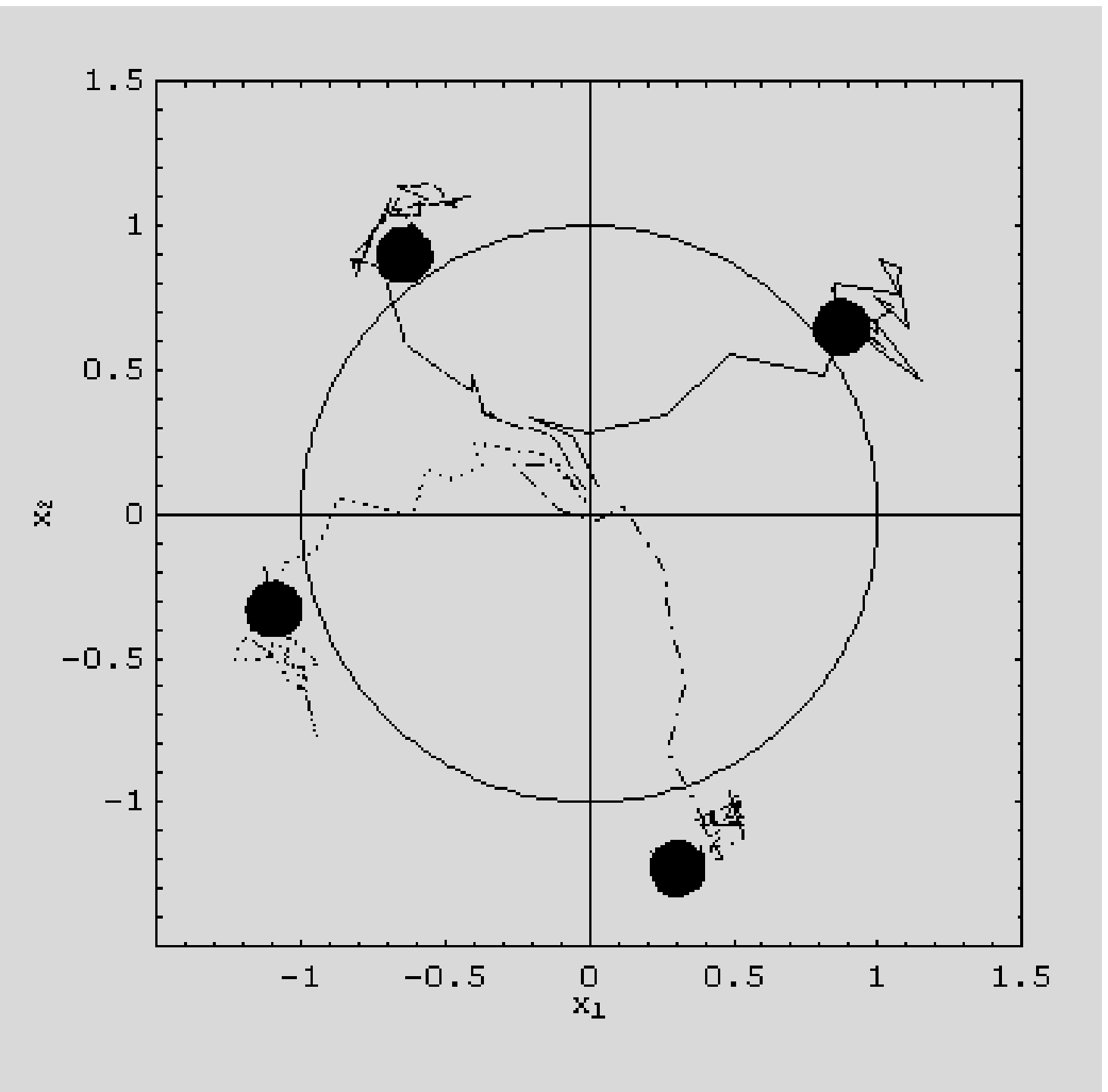}
\par\end{center}
\begin{itemize}
\item $M=4$ and $n=10$ were used.
\item The reference vectors $\boldsymbol{x}^{\prime}\left(y\right)$ (for
$y=1,2,3,4$) are initialised close to the origin.
\item The training history leads to stationary $\boldsymbol{x}^{\prime}\left(y\right)$
just outside the unit circle.
\end{itemize}

\subsubsection{Posterior probabilities}

\begin{center}
\includegraphics[width=8cm]{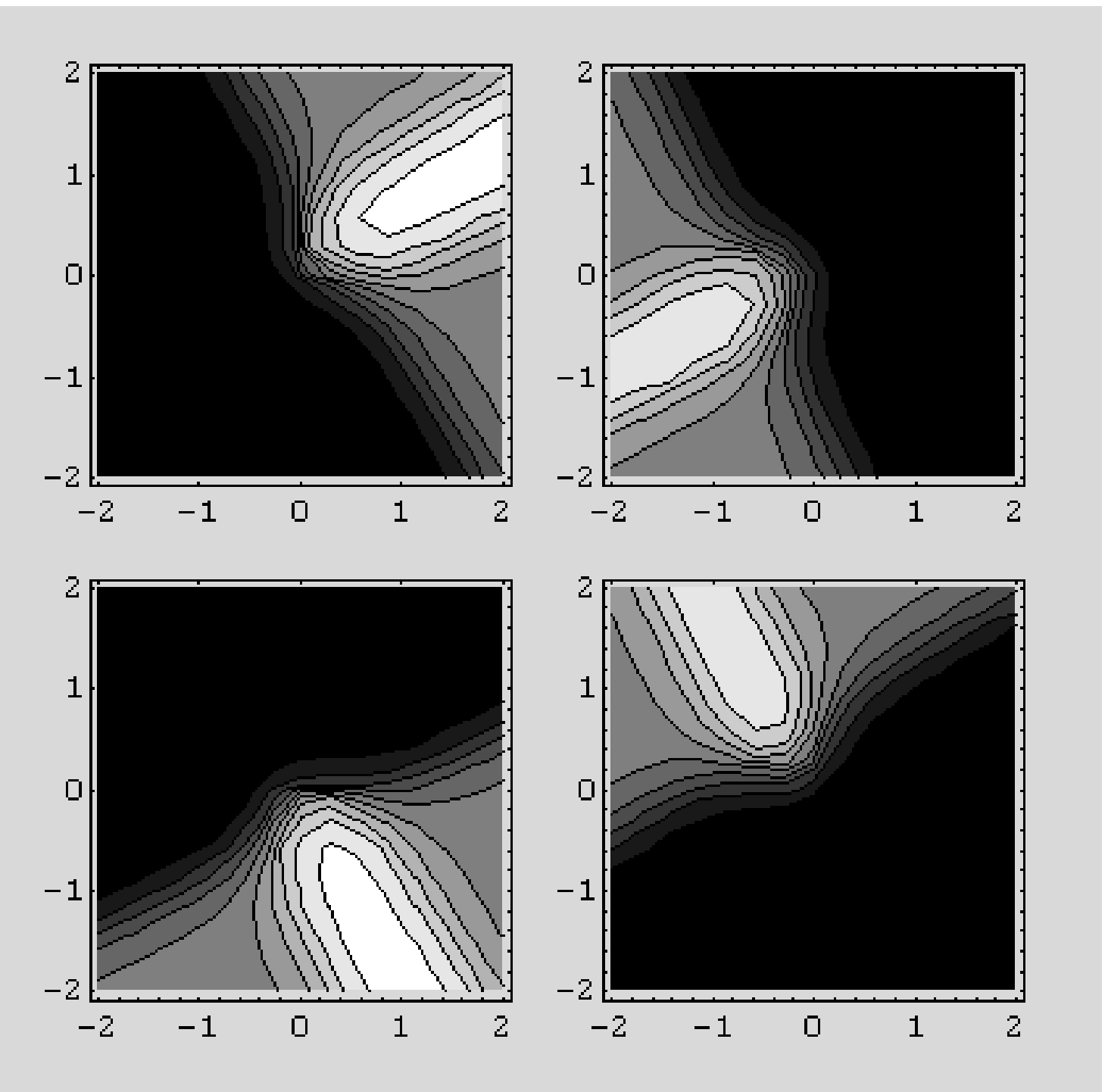}
\par\end{center}
\begin{itemize}
\item Each of the posterior probabilities $\Pr\left(y\left|\boldsymbol{x}\right.\right)$
(for $y=1,2,3,4$) is large mainly in a $\frac{\pi}{2}$ radian arc
of the circle.
\item There is some overlap between the $\Pr\left(y\left|\boldsymbol{x}\right.\right)$.
\end{itemize}

\subsection{2-Torus}

\subsubsection{Posterior probabilities: joint encoding}

\begin{center}
\includegraphics[width=8cm]{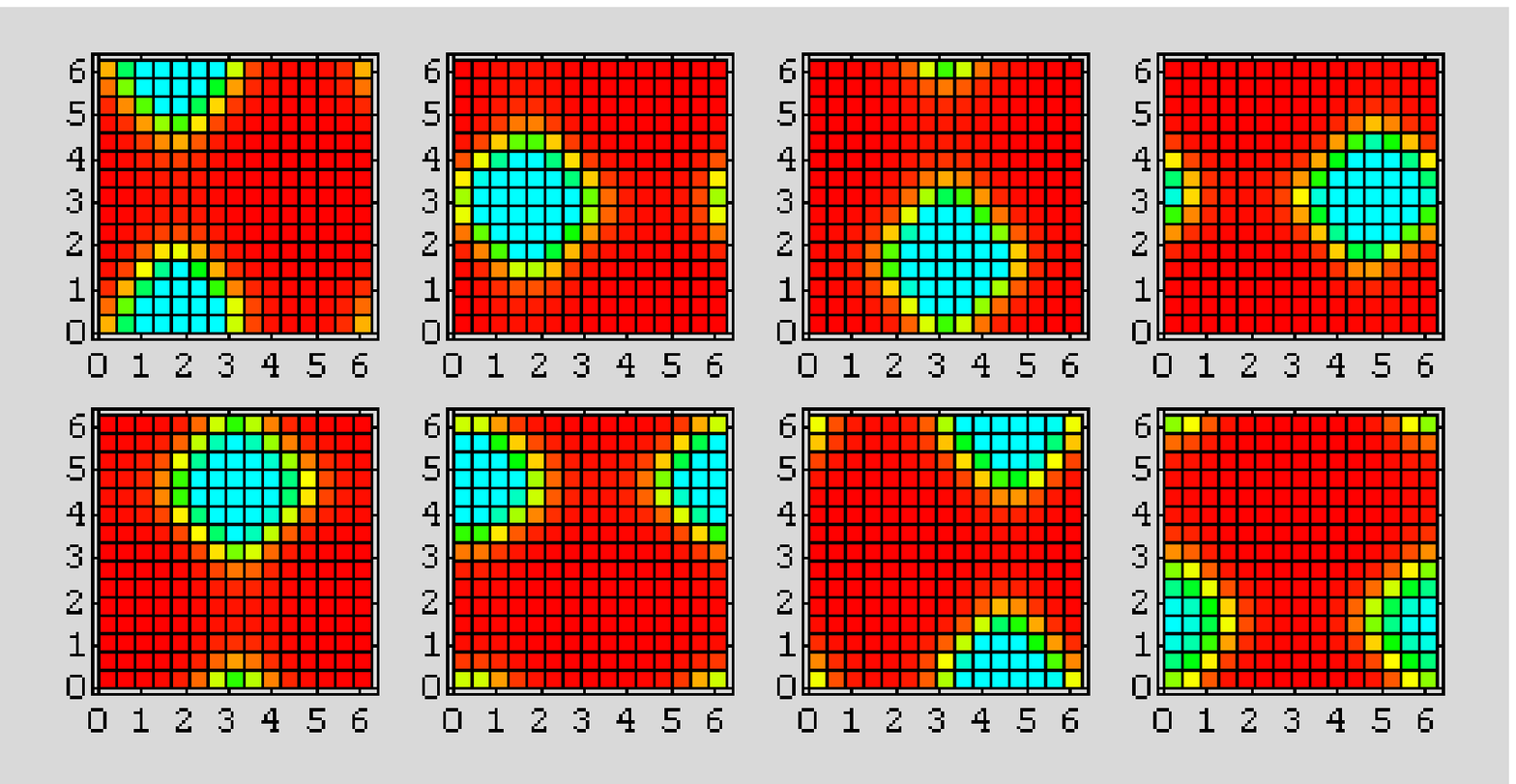}
\par\end{center}
\begin{itemize}
\item $M=8$ and $n=5$ were used, which lies inside the joint encoding
region of the stability diagram.
\item Each of the posterior probabilities $\Pr\left(y\left|\boldsymbol{x}\right.\right)$
is large mainly in a localised region of the torus.
\item There is some overlap between the $\Pr\left(y\left|\boldsymbol{x}\right.\right)$.
\end{itemize}

\subsubsection{Posterior probabilities: factorial encoding}

\begin{center}
\includegraphics[width=8cm]{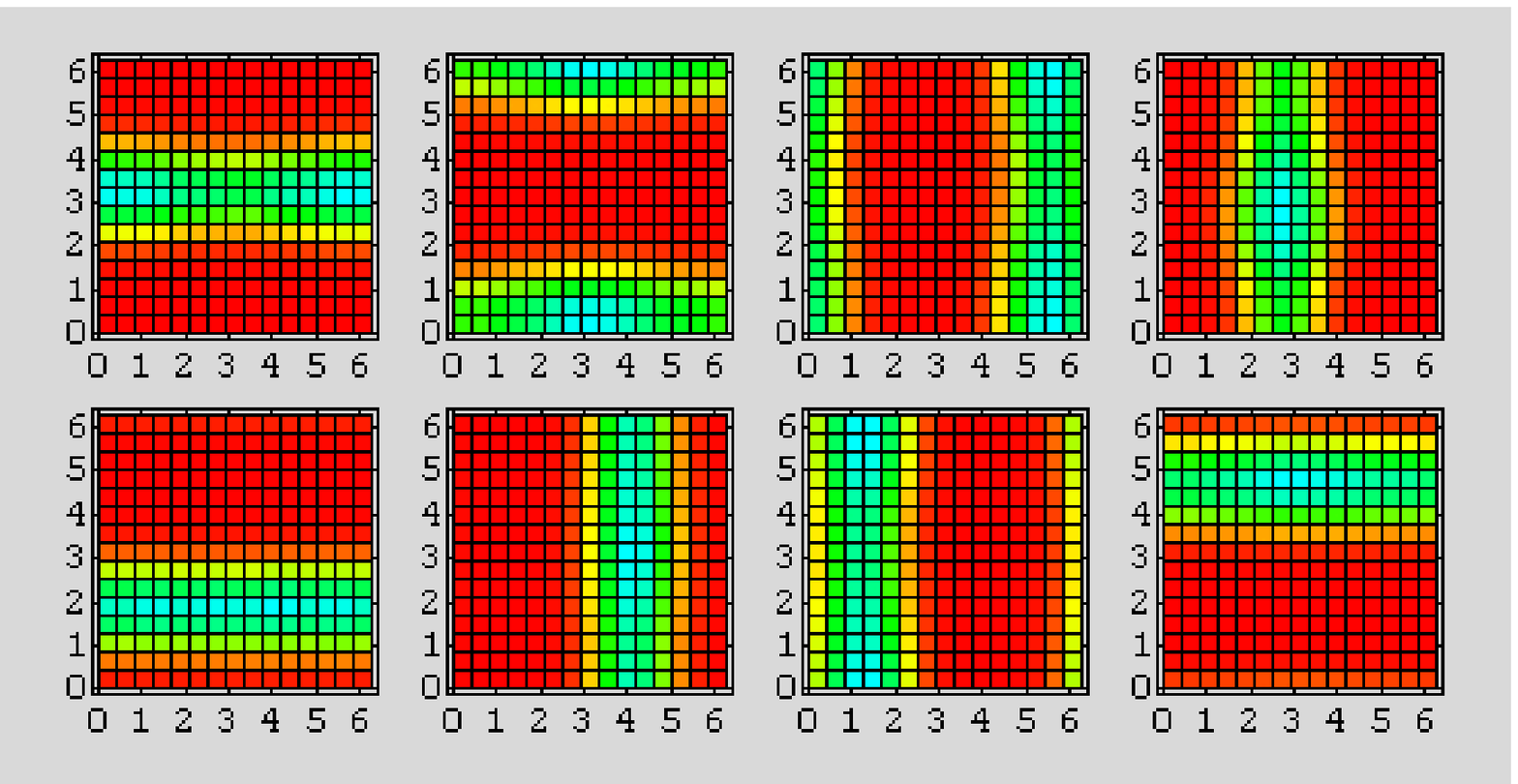}
\par\end{center}
\begin{itemize}
\item $M=8$ and $n=20$ were used, which lies inside the factorial encoding
region of the stability diagram.
\item Each of the posterior probabilities $\Pr\left(y\left|\boldsymbol{x}\right.\right)$
is large mainly in a collar-shaped region of the torus; half circle
one way round the torus, and half the other way.
\item There is some overlap between the $\Pr\left(y\left|\boldsymbol{x}\right.\right)$
that circle the same way round the torus.
\item There is a localised region of overlap between a pair of $\Pr\left(y\left|\boldsymbol{x}\right.\right)$
that circle the opposite way round the torus.
\item These localised overlap regions are the mechanism by which factorial
encoding has a small reconstruction distortion.
\end{itemize}

\subsection{Multiple Independent Targets}

\subsubsection{Training data}

\begin{center}
\includegraphics[width=8cm]{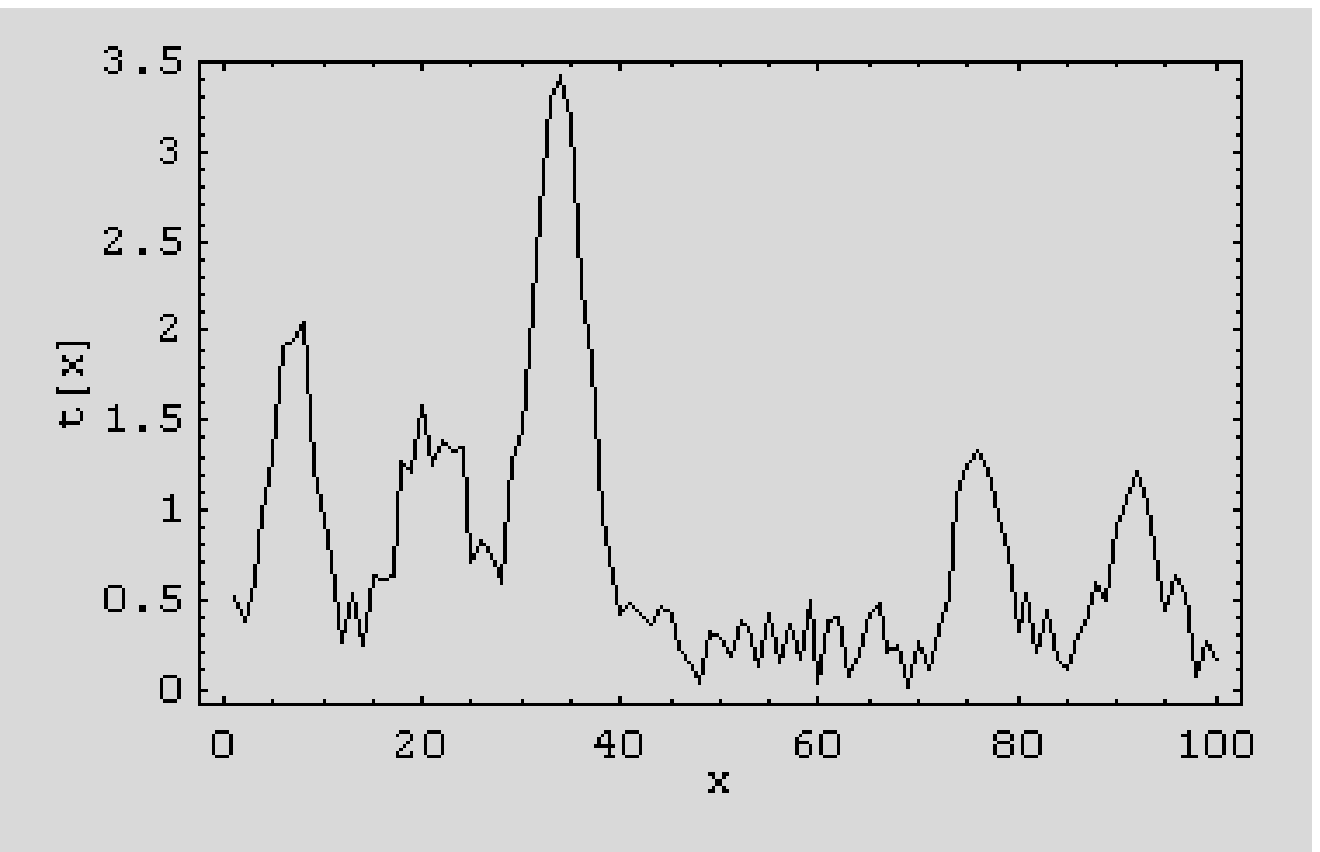}
\par\end{center}
\begin{itemize}
\item The targets were unit height Gaussian bumps with $\sigma=2$.
\item The additive noise was uniformly distributed variables in $\left[0,0.5\right]$.
\end{itemize}

\subsubsection{Factorial encoding}

\begin{center}
\includegraphics[width=8cm]{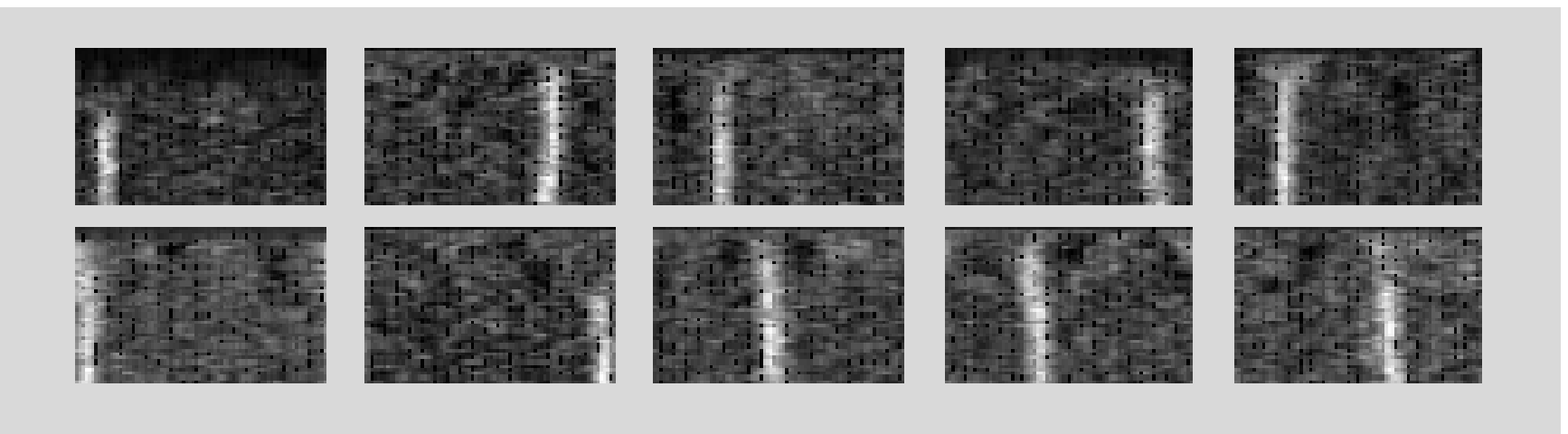}
\par\end{center}
\begin{itemize}
\item $M=10$ and $n=10$ were used.
\item Each of the reference vectors $\boldsymbol{x}^{\prime}\left(y\right)$
becomes large in a localised region.
\item Each input vector causes a subset of the neurons to fire corresponding
to locations of the targets.
\item This is a factorial encoder because each neuron responds to only a
subspace of the input.
\end{itemize}

\subsection{Pair of Correlated Targets}

\subsubsection{Training data}

\begin{center}
\includegraphics[width=8cm]{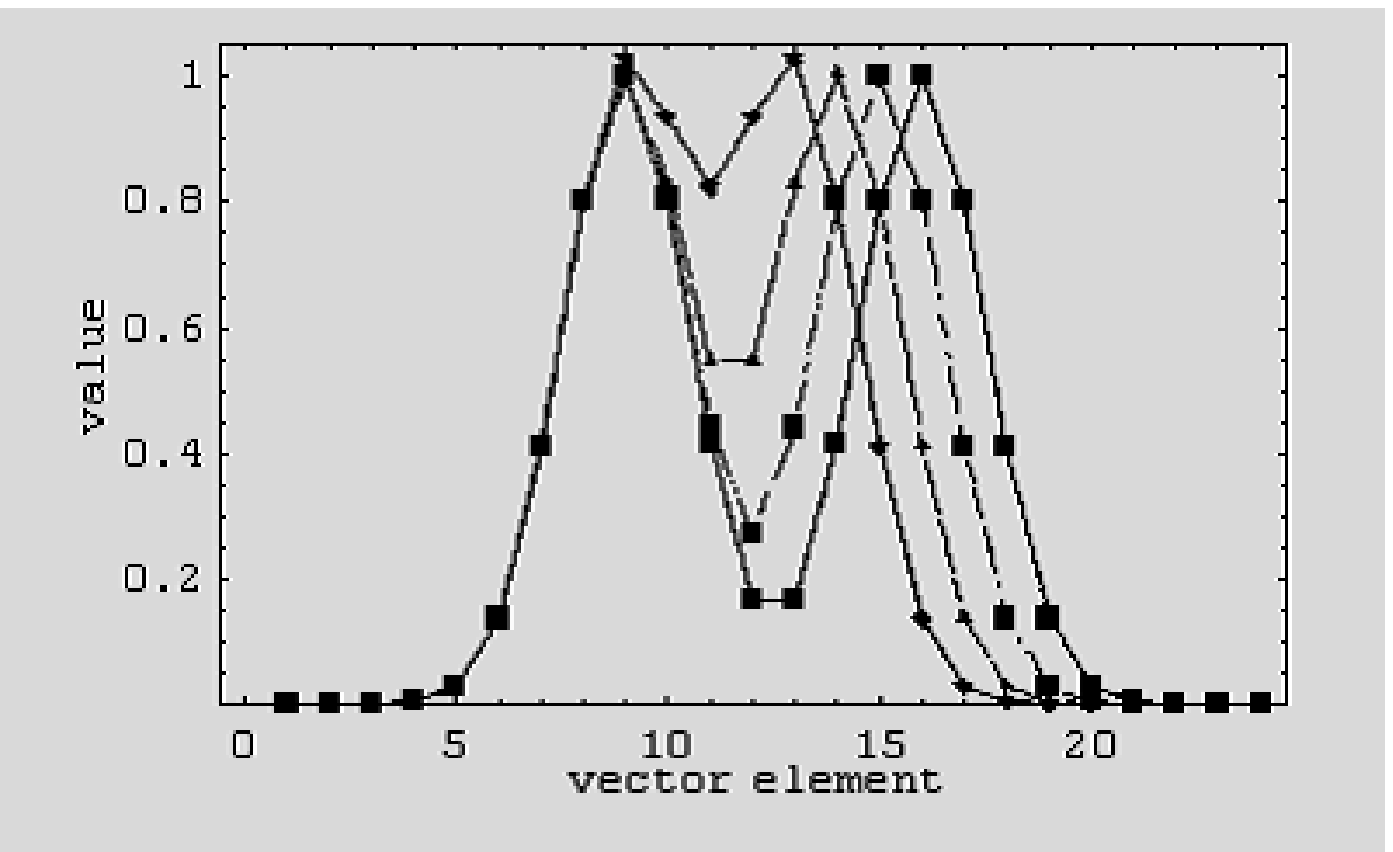}
\par\end{center}
\begin{itemize}
\item The targets were unit height Gaussian bumps with $\sigma=1.5$.
\end{itemize}

\subsubsection{Training history: joint encoding}

\begin{center}
\includegraphics[width=8cm]{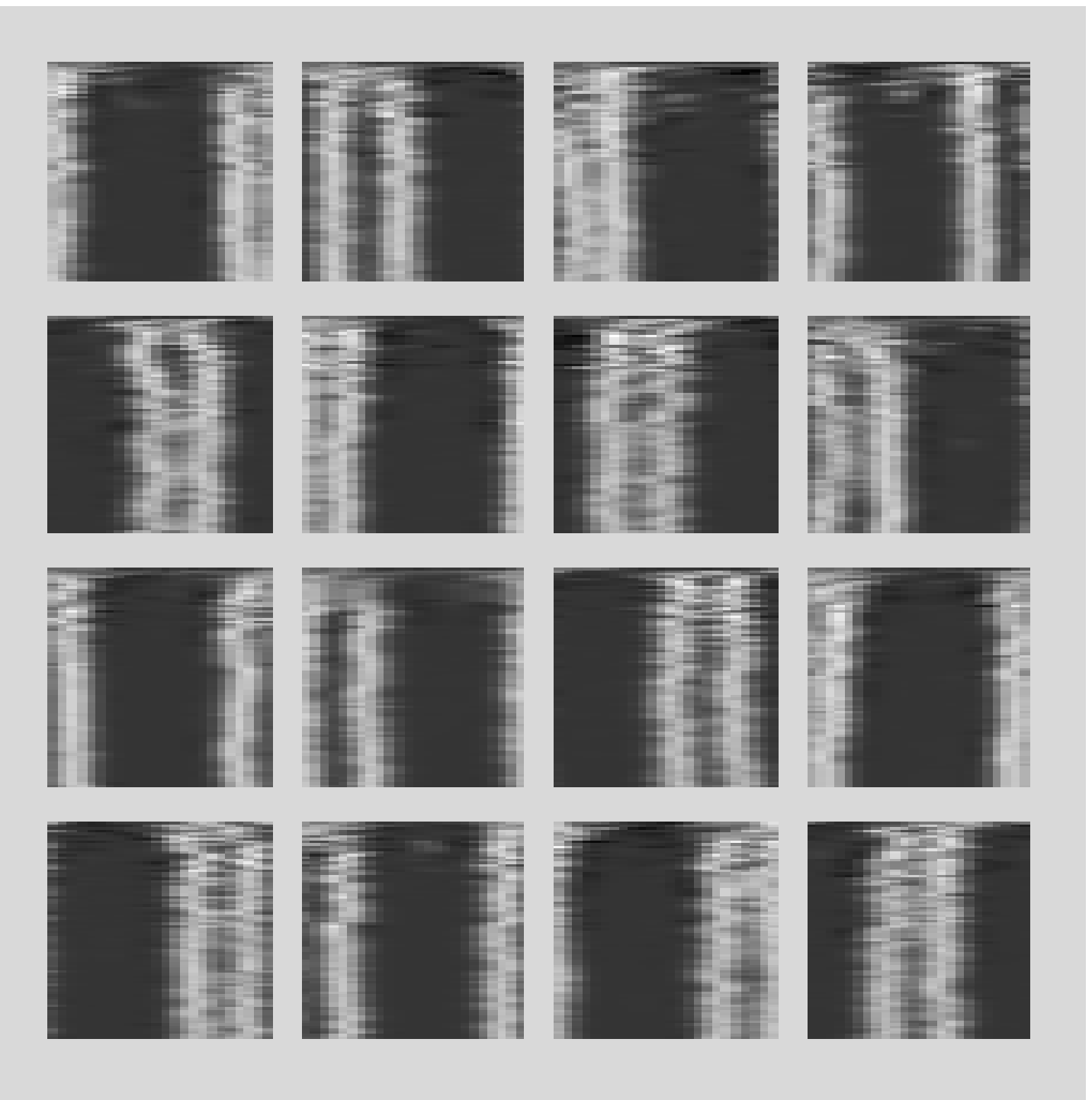}
\par\end{center}
\begin{itemize}
\item $M=16$ and $n=3$ were used.
\item Each of the reference vectors $\boldsymbol{x}^{\prime}\left(y\right)$
becomes large in a pair of localised regions.
\item Each neuron responds to a small range of positions and separations
of the pair of targets.
\item The neurons respond \emph{jointly} to the position and separation
of the targets.
\end{itemize}

\subsubsection{Training history: factorial encoding}

\begin{center}
\includegraphics[width=8cm]{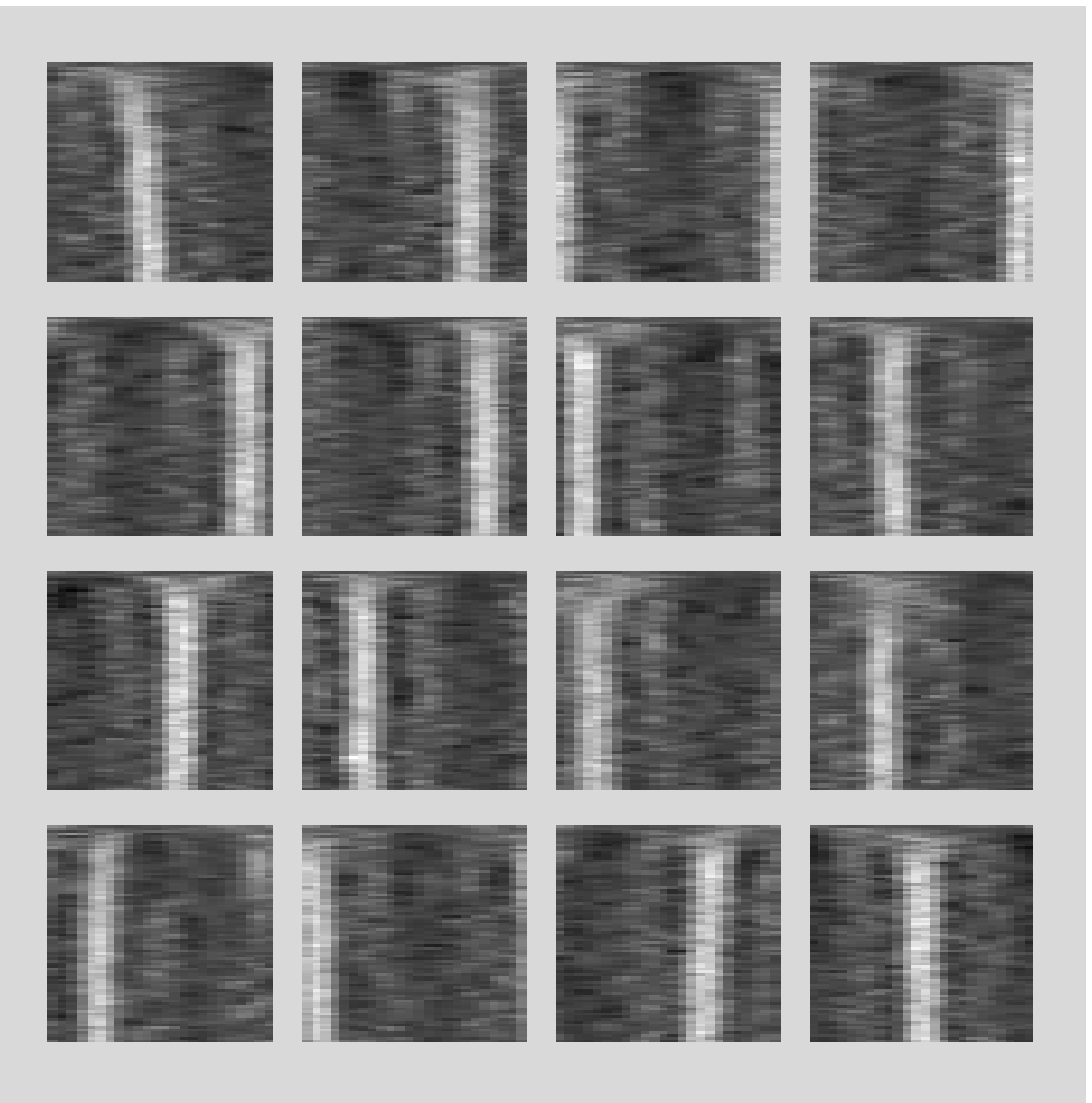}
\par\end{center}
\begin{itemize}
\item $M=\{16,16\}$ and $n=\{20,20\}$ were used; this is a 2-stage encoder.
\item The second encoder uses as input the posterior probability output
by the first encoder.
\item The objective function is the sum of the separate encoder objective
functions (with equal weighting given to each).
\item The presence of the second encoder affects the optimisation of the
first encoder via {}``self-supervision''.
\item Each of the reference vectors $\boldsymbol{x}^{\prime}\left(y\right)$
becomes large in a single localised region.
\end{itemize}

\subsubsection{Training history: invariant encoding}

\begin{center}
\includegraphics[width=8cm]{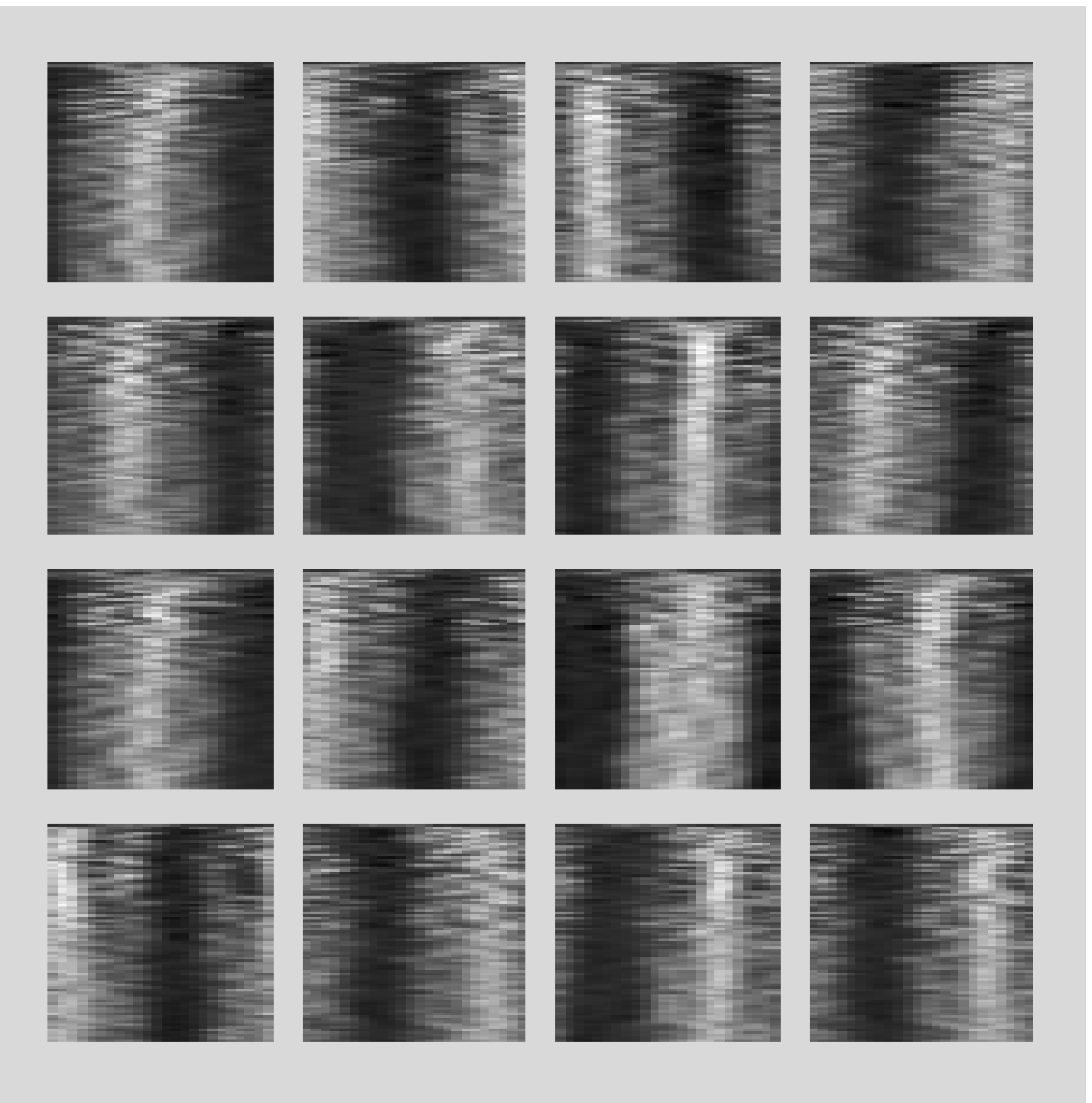}
\par\end{center}
\begin{itemize}
\item $M=\{16,16\}$ and $n=\{3,3\}$ were used; this is a 2-stage encoder.
\item During training the ratio of the weighting assigned to the first and
second encoders is increased from 1:5 to 1:40.
\item Each of the reference vectors $\boldsymbol{x}^{\prime}\left(y\right)$
becomes large in a single broad region.
\item Each neuron responds only the position (and not the separation) of
the pair of targets.
\item The response of the neurons is \emph{invariant} w.r.t. the separation
of the targets.
\end{itemize}

\subsection{Separating Different Waveforms}

\subsubsection{Training data}

\begin{center}
\includegraphics[width=8cm]{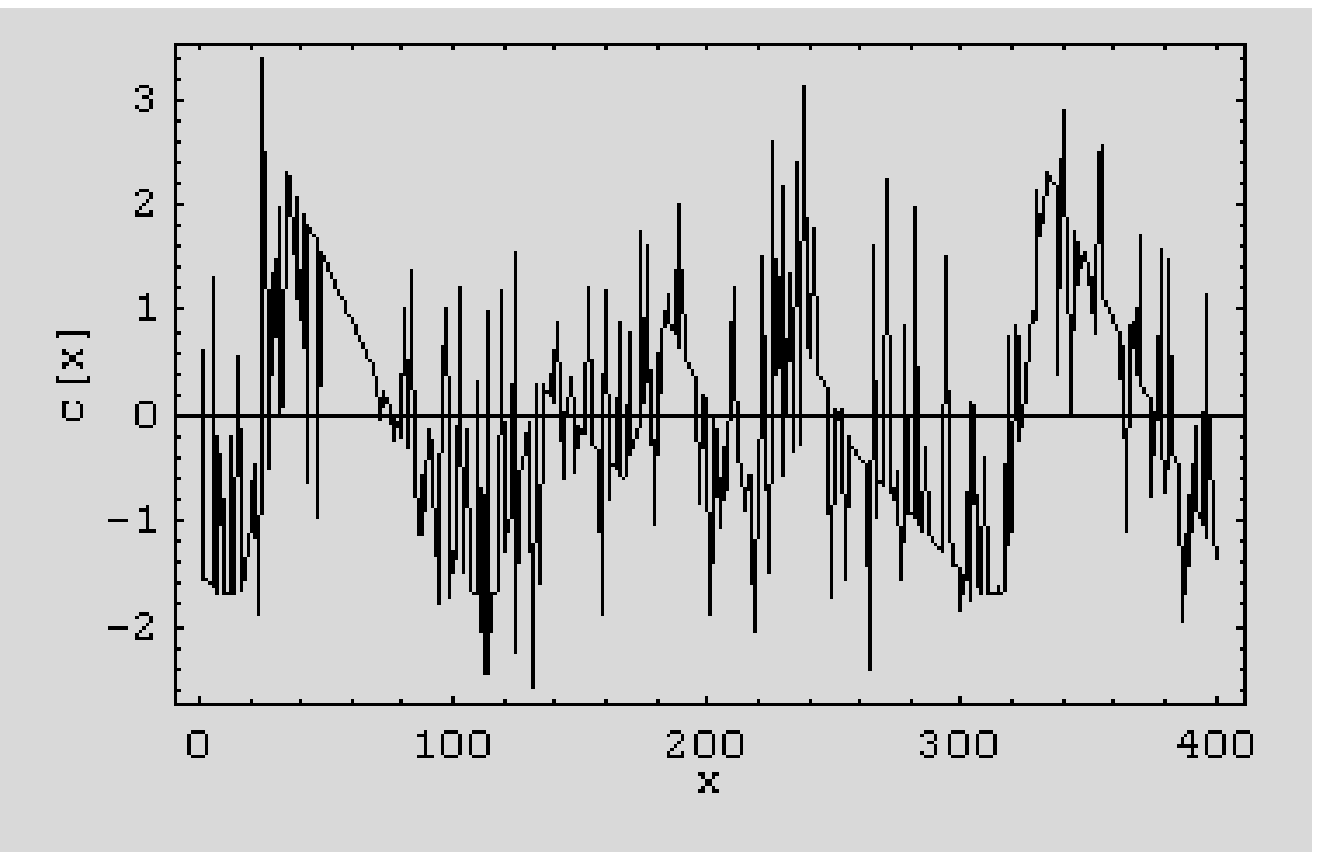}
\par\end{center}
\begin{itemize}
\item This data is the superposition of a pair of waveforms plus noise.
\item In each training vector the relative phase of the two waveforms is
randomly selected.
\end{itemize}

\subsubsection{Training history: factorial encoding}

\begin{center}
\includegraphics[width=8cm]{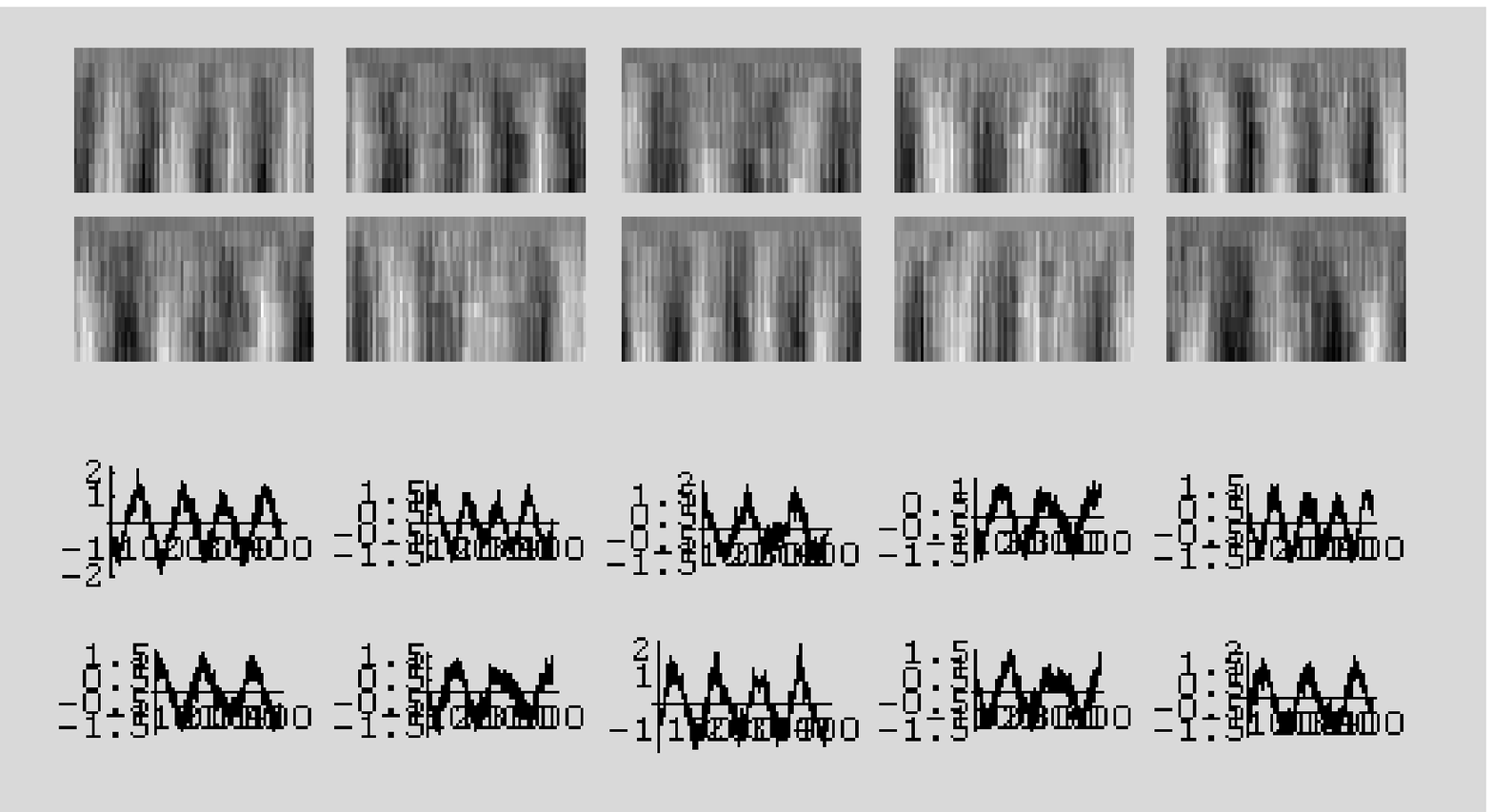}
\par\end{center}
\begin{itemize}
\item $M=10$ and $n=20$ were used.
\item Each of the reference vectors $\boldsymbol{x}^{\prime}\left(y\right)$
becomes one or other of the two waveforms, and has a definite phase.
\item Each neuron responds to only one of the waveforms, and then only when
its phase is in a localised range.
\end{itemize}

\subsection{Maternal + Foetal ECG}

\subsubsection{Training data}

\begin{center}
\includegraphics[width=8cm]{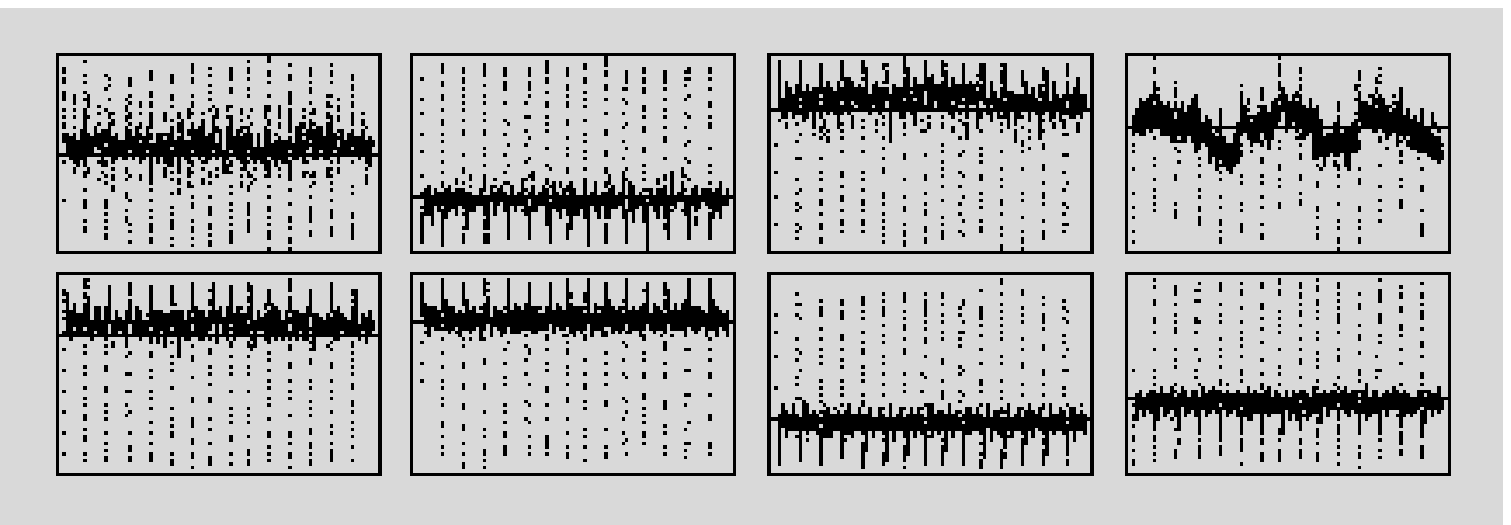}
\par\end{center}
\begin{itemize}
\item This data is an 8-channel ECG recording taken from a pregnant woman.
\item The large spikes are the woman's heart beat.
\item The noise masks the foetus' heartbeat.
\item This data was whitened before training the neural network.
\end{itemize}

\subsubsection{Factorial Encoding}

\begin{center}
\includegraphics[width=8cm]{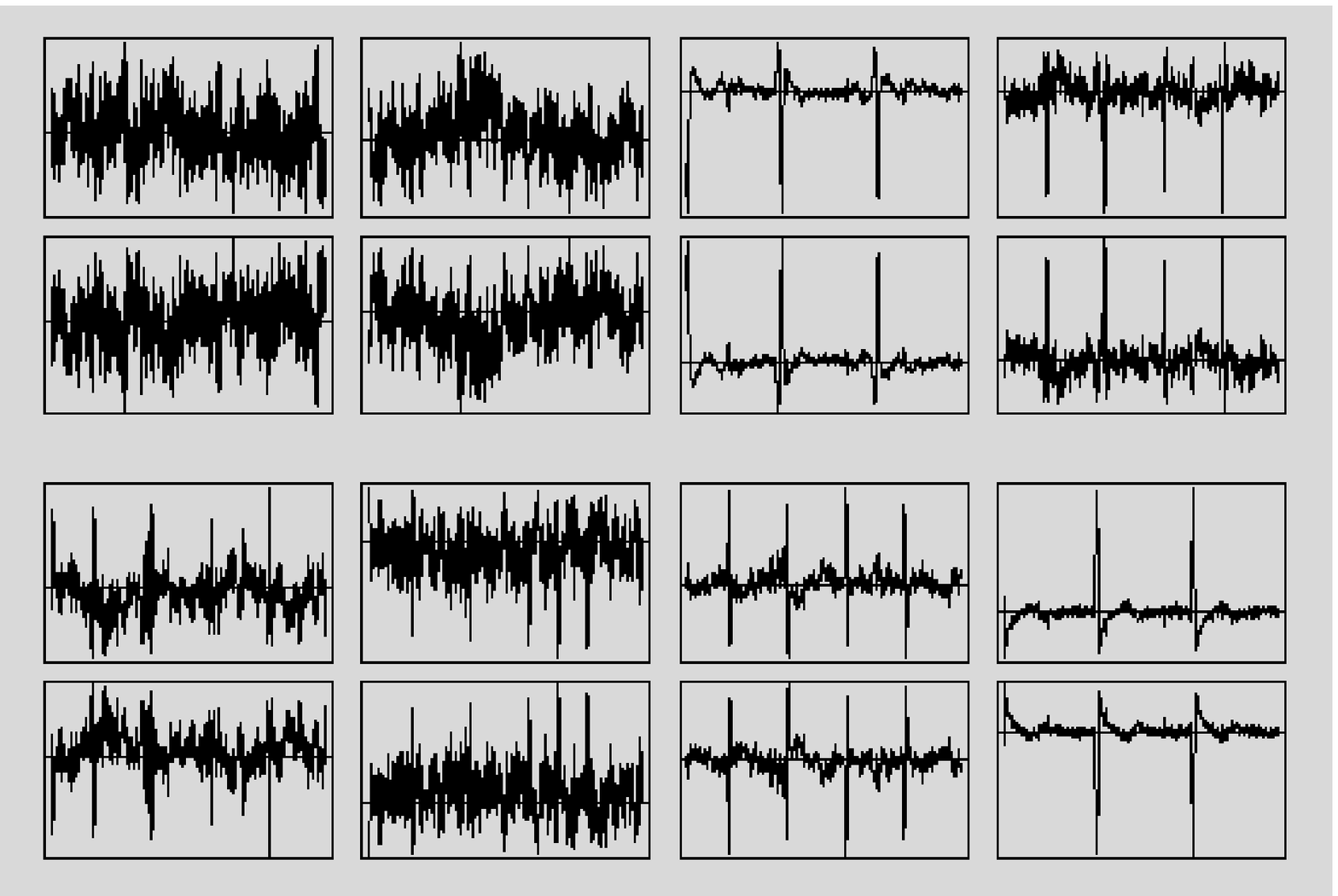}
\par\end{center}
\begin{itemize}
\item $M=16$ and $n=20$ were used.
\item The results shown are $\boldsymbol{w}(y).\boldsymbol{x}$ computed
for all neurons ($y=1,2,\,\cdots\,,8$) for each 8-dimensional input
vector $\boldsymbol{x}$.
\item After limited training some, but not all, of the neurons have converged.
\item The broadly separated spikes indicate a neuron that responds to the
mother's heartbeat.
\item The closely separated spikes indicate a neuron that responds to the
foetus' heartbeat.
\end{itemize}

\subsection{Visual Cortex Network (VICON)}

\subsubsection{Training Data}

\begin{center}
\includegraphics[width=8cm]{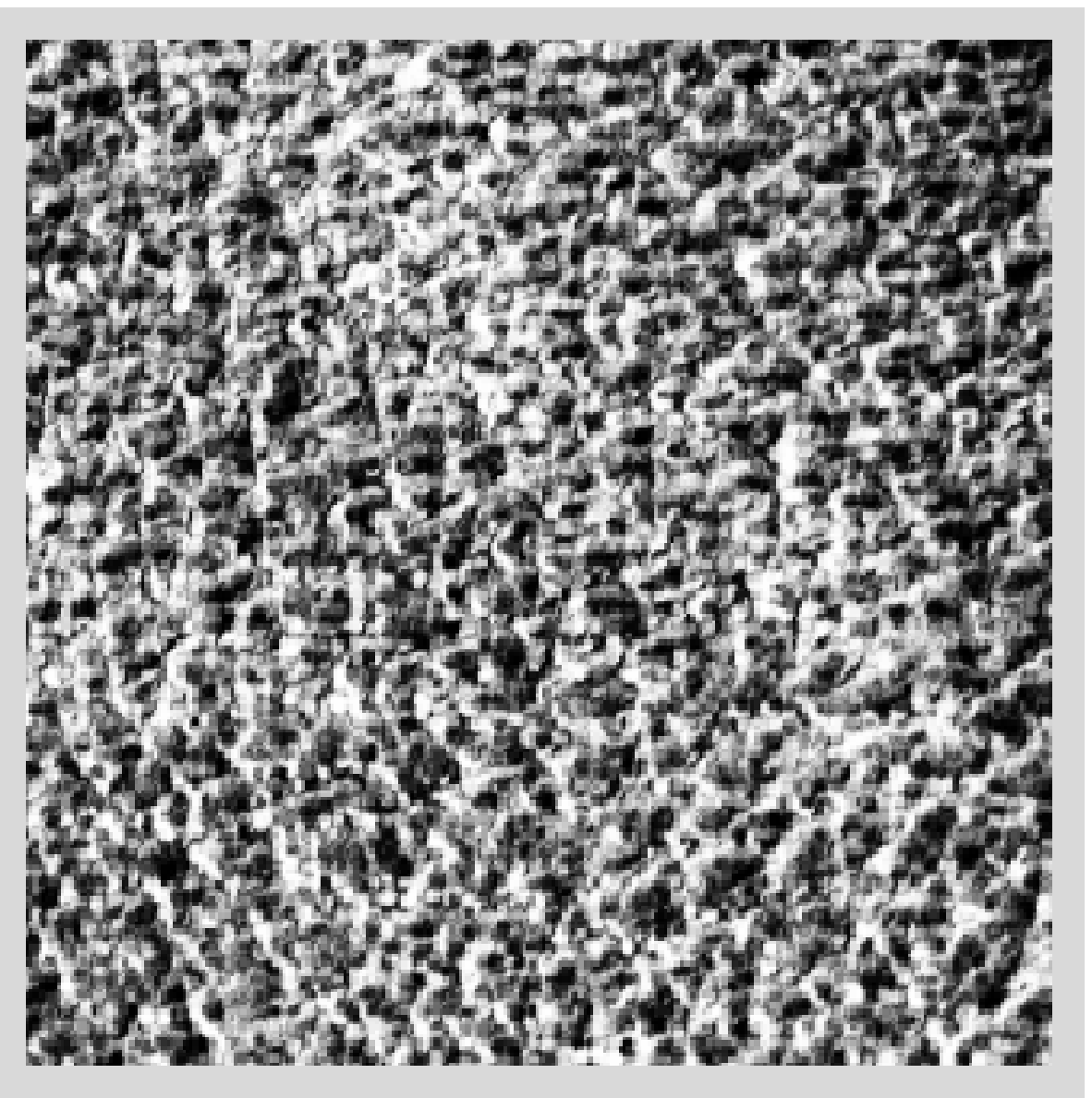}
\par\end{center}
\begin{itemize}
\item This is a Brodatz texture image, whose spatial correlation length
is 5-10 pixels.
\end{itemize}

\subsubsection{Orientation map}

\begin{center}
\includegraphics[width=8cm]{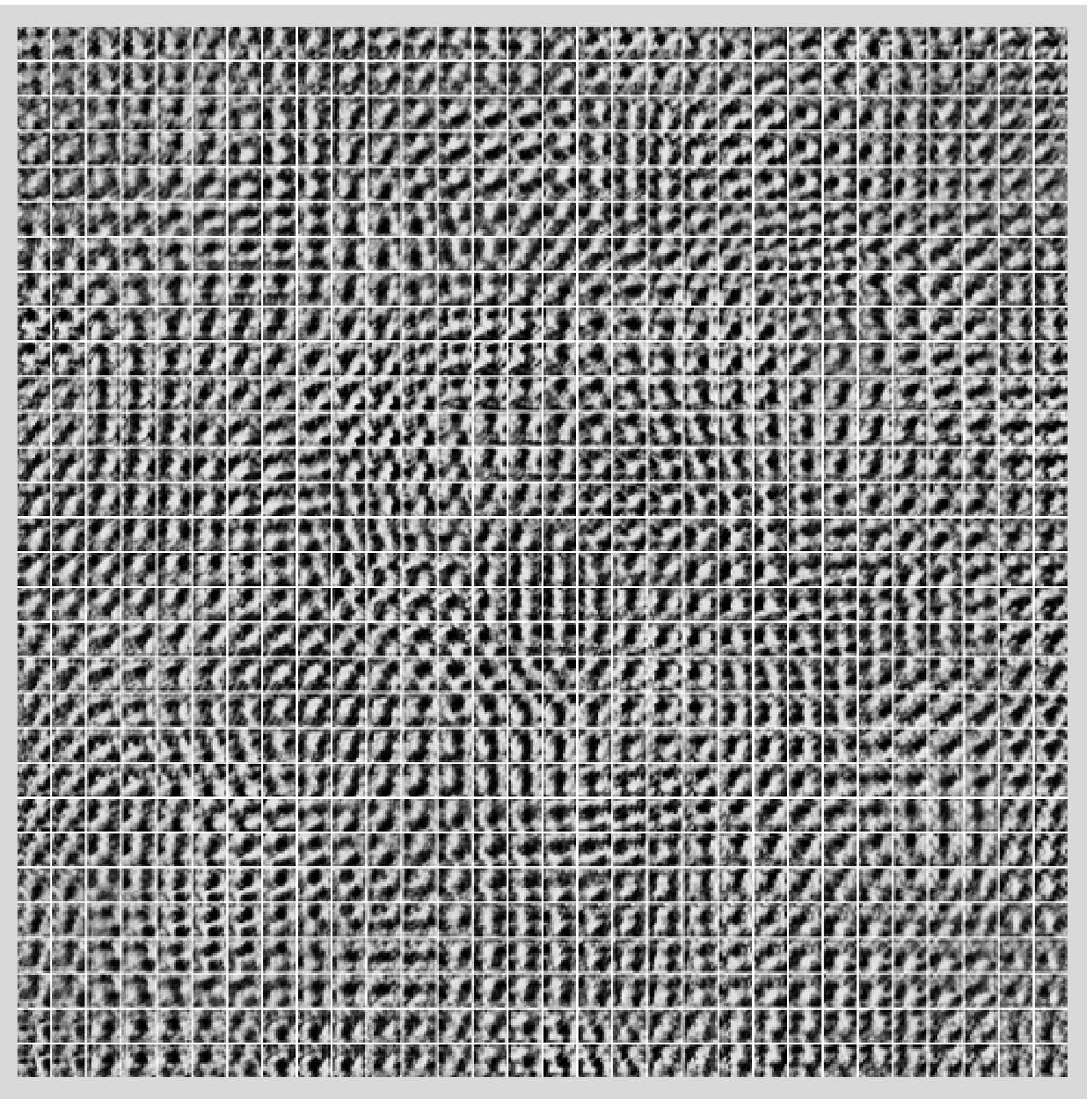}
\par\end{center}
\begin{itemize}
\item $M=30\times30$ and $n=1$ were used.
\item Input window size = $17\times17$, neighbourhood size = $9\times9$,
leakage neighbourhood size = $3\times3$ were used.
\item Leakage probability was sampled from a 2-dimensional Gaussian PDF,
with $\sigma=1$ in each direction.
\item Each of the reference vectors $\boldsymbol{x}^{\prime}\left(y\right)$
typically looks like a small patch of image.
\item Leakage induces topographic ordering across the array of neurons
\item This makes the array of reference vectors look like an {}``orientation
map''.
\end{itemize}

\subsubsection{Sparse coding}

\begin{center}
\includegraphics[width=8cm]{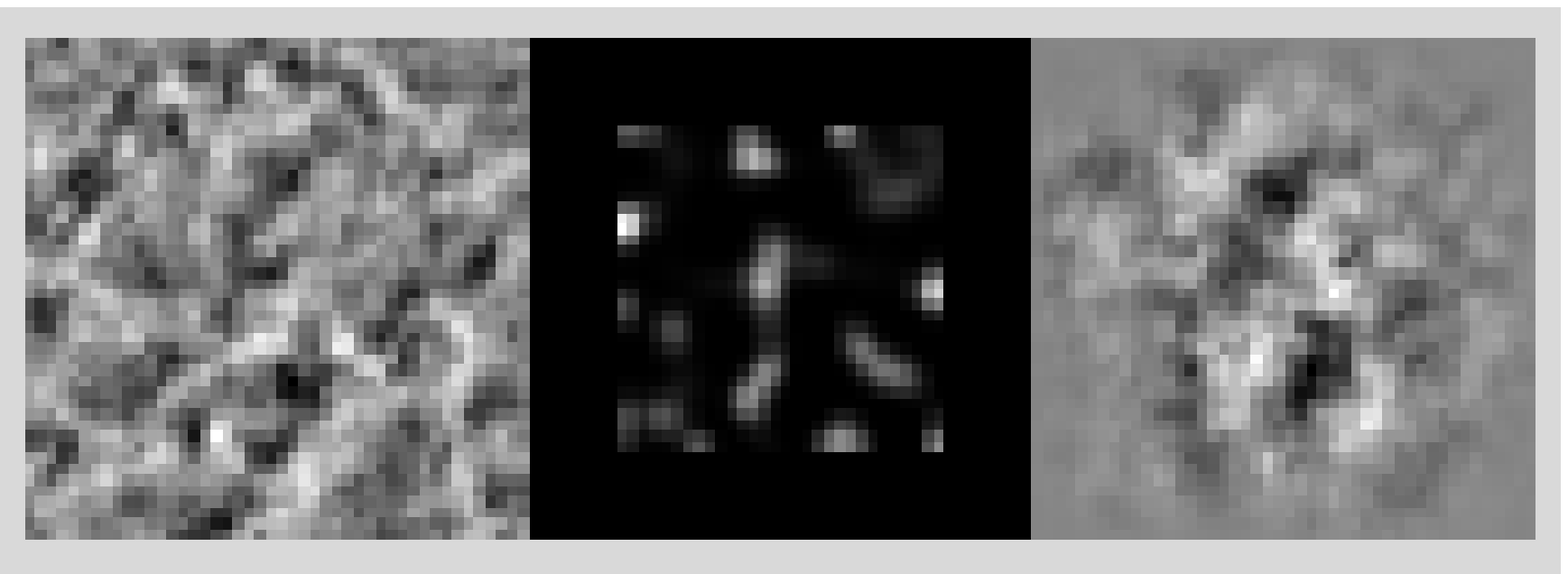}
\par\end{center}
\begin{itemize}
\item The trained network is used to encode and decode a typical input image.
\item Left image = input.
\item Middle image = posterior probability. This shows {}``sparse coding''
with a small number of {}``activity bubbles''.
\item Right image = reconstruction. Apart from edge effects, this is a low
resolution version of the input.
\end{itemize}

\subsubsection{Dominance stripes}

\begin{center}
\includegraphics[width=8cm]{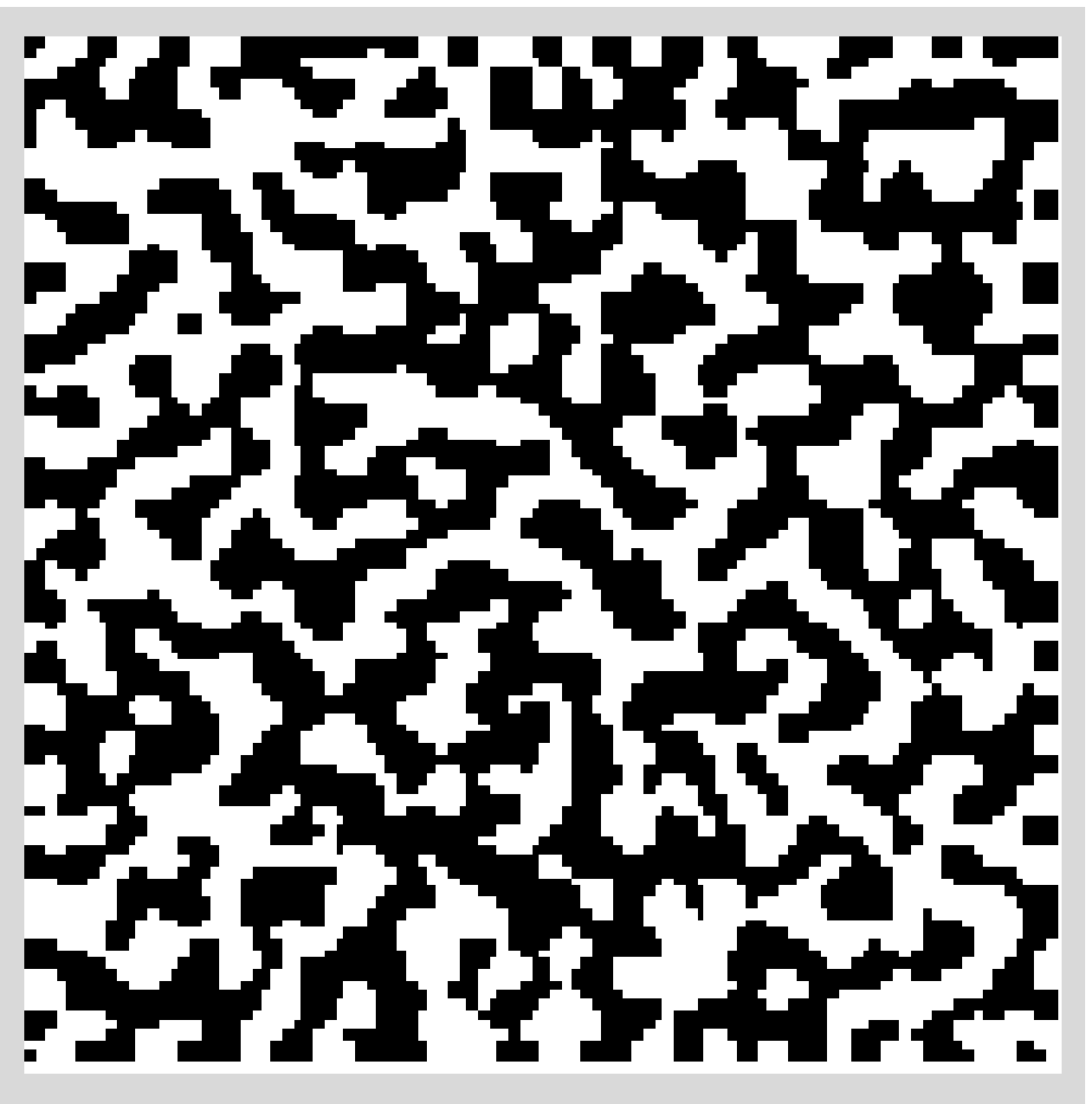}
\par\end{center}
\begin{itemize}
\item Interdigitate a pair of training images, so that one occupies on the
black squares, and the other the white squares, of a {}``chess board''.
\item Preprocess this interdigitated image to locally normalise it using
a finite range neighbourhood.
\item $M=100\times100$ and $n=1$ were used.
\item Input window size = $3\times3$, neighbourhood size = $5\times5$,
leakage neighbourhood size = $3\times3$ were used.
\item Leakage probability was sampled from a 2-dimensional Gaussian PDF,
with $\sigma=1$ in each direction.
\item The dominance stripe map records for each neuron which of the 2 interdigitated
images causes it to respond more strongly.
\item The dominance stripes tend to run perpendicularly into the boundaries,
because the neighbourhood window is truncated at the edge of the array.
\end{itemize}

\section{References}

\noindent Luttrell S P, 1997, to appear in \emph{Proceedings of the
Conference on Information Theory and the Brain}, Newquay, 20-21 September
1996, The emergence of dominance stripes and orientation maps in a
network of firing neurons.\\

\noindent Luttrell S P, 1997, \emph{Mathematics of Neural Networks:
Models, Algorithms and Applications}, Kluwer, Ellacott S W, Mason
J C and Anderson I J (eds.), A theory of self-organising neural networks,
240-244.\\

\noindent Luttrell S P, 1999, \emph{Combining Artificial Neural Nets:
Ensemble and Modular Multi-Net Systems}, 235-263, Springer-Verlag,
Sharkey A J C (ed.), Self-organised modular neural networks for encoding
data. \\

\noindent Luttrell S P, 1999, An Adaptive Network For Encoding Data
Using Piecewise Linear Functions, \emph{Proceedings of the 9th International
Conference on Artificial Neural Networks (ICANN99)}, Edinburgh, 7-10
September 1999, 198-203.\\

\noindent Luttrell S P, 1999, submitted to a special issue of \emph{IEEE
Trans. Information Theory on Information-Theoretic Imaging}, Stochastic
vector quantisers.
\end{document}